\documentclass[letterpaper]{article} 
\usepackage{aaai2026}  
\usepackage{times}  
\usepackage{helvet}  
\usepackage{courier}  
\usepackage[hyphens]{url}  
\usepackage{graphicx} 
\usepackage{natbib}  
\usepackage{caption} 
\usepackage{algorithm}
\usepackage{algorithmic}

\usepackage{amssymb}
\usepackage{enumitem}
\usepackage{booktabs}
\usepackage{subcaption}
\usepackage{array}
\usepackage{fancyvrb}
\usepackage{fancyhdr}
\usepackage{xcolor}
\usepackage[most]{tcolorbox}
\usepackage{parskip}
\usepackage{url}

\definecolor{lowyellow}{RGB}{241, 196, 15}

\newcommand{\mn}{MATPO }
\newcommand{\modelname}{Multi-Agent Tool-Integrated Policy Optimization }

\usepackage{newfloat}
\usepackage{listings}
\DeclareCaptionStyle{ruled}{labelfont=normalfont,labelsep=colon,strut=off} 
\floatstyle{ruled}
\newfloat{listing}{tb}{lst}{}
\floatname{listing}{Listing}
\title{Multi-Agent Tool-Integrated Policy Optimization}
\author{%
  Zhanfeng Mo\thanks{\; Equal contribution.},
  Xingxuan Li\footnotemark[1],
  Yuntao Chen,
  Lidong Bing\thanks{\; Corresponding author.}
}
\affiliations{
    MiroMind AI\\
    \{zhanfeng.mo,xingxuan.li,yuntao.chen,lidong.bing\}@miromind.ai

}

\usepackage{bibentry}

\begin{document}

\maketitle

\begin{abstract}
    Large language models (LLMs) increasingly rely on multi-turn tool-integrated planning for knowledge-intensive and complex reasoning tasks.
    Existing implementations typically rely on a single agent, but they suffer from limited context length and noisy tool responses.
    A natural solution is to adopt a multi-agent framework with planner- and worker-agents to manage context. 
    However, no existing methods support effective reinforcement learning post-training of tool-integrated multi-agent frameworks. 
    To address this gap, we propose \modelname (\textbf{MATPO}), which enables distinct roles (planner and worker) to be trained within a single LLM instance using role-specific prompts via reinforcement learning.
    \mn is derived from a principled credit assignment mechanism across planner and worker rollouts. This design eliminates the need to deploy multiple LLMs, which would be memory-intensive, while preserving the benefits of specialization.
    Experiments on \texttt{GAIA-text}, \texttt{WebWalkerQA}, and \texttt{FRAMES} show that \mn consistently outperforms single-agent baselines by an average of $18.38\%$ relative improvement in performance and exhibits greater robustness to noisy tool outputs. Our findings highlight the effectiveness of unifying multiple agent roles within a single LLM and provide practical insights for stable and efficient multi-agent RL training. 
    \footnote{Our code is available at \url{https://github.com/mzf666/MATPO}.}
\end{abstract}

\section{Introduction}
\label{sec:introduction}
Advancements in AI agent capabilities increasingly rely on sophisticated multi-turn tool-integrated planning (TIP) \citep{toolstar, toolrl}, where large language models (LLMs) iteratively perform planning and leverage specialized tools, such as search tools for information retrieval, coding tools for analysis, and file-reading tools for document processing.
Among these tools, the search tool has emerged as particularly crucial, allowing LLMs to access external information that extends far beyond their parametric knowledge to support in-depth investigation and analysis.

Current implementations typically enable a single agent to conduct deep research \citep{arpo, searchr1} through iterative multi-turn interactions with search tools, allowing the agent to progressively gather, analyze, and summarize information from multiple sources. However, this single-agent approach faces several significant limitations that hinder its effectiveness in complex, real-world research scenarios: 
1. tool-responses (\textit{e.g.,} searching or scraping websites) often consume a large number of tokens, making long-range multi-turn TIP prohibitive under the LLM’s limited context length; 
2. tool-responses are often noisy and can interfere with the LLM’s attention and planning, hindering its ability to plan high-quality subsequent actions.

\begin{figure}
    \centering
    \includegraphics[width=0.9\linewidth]{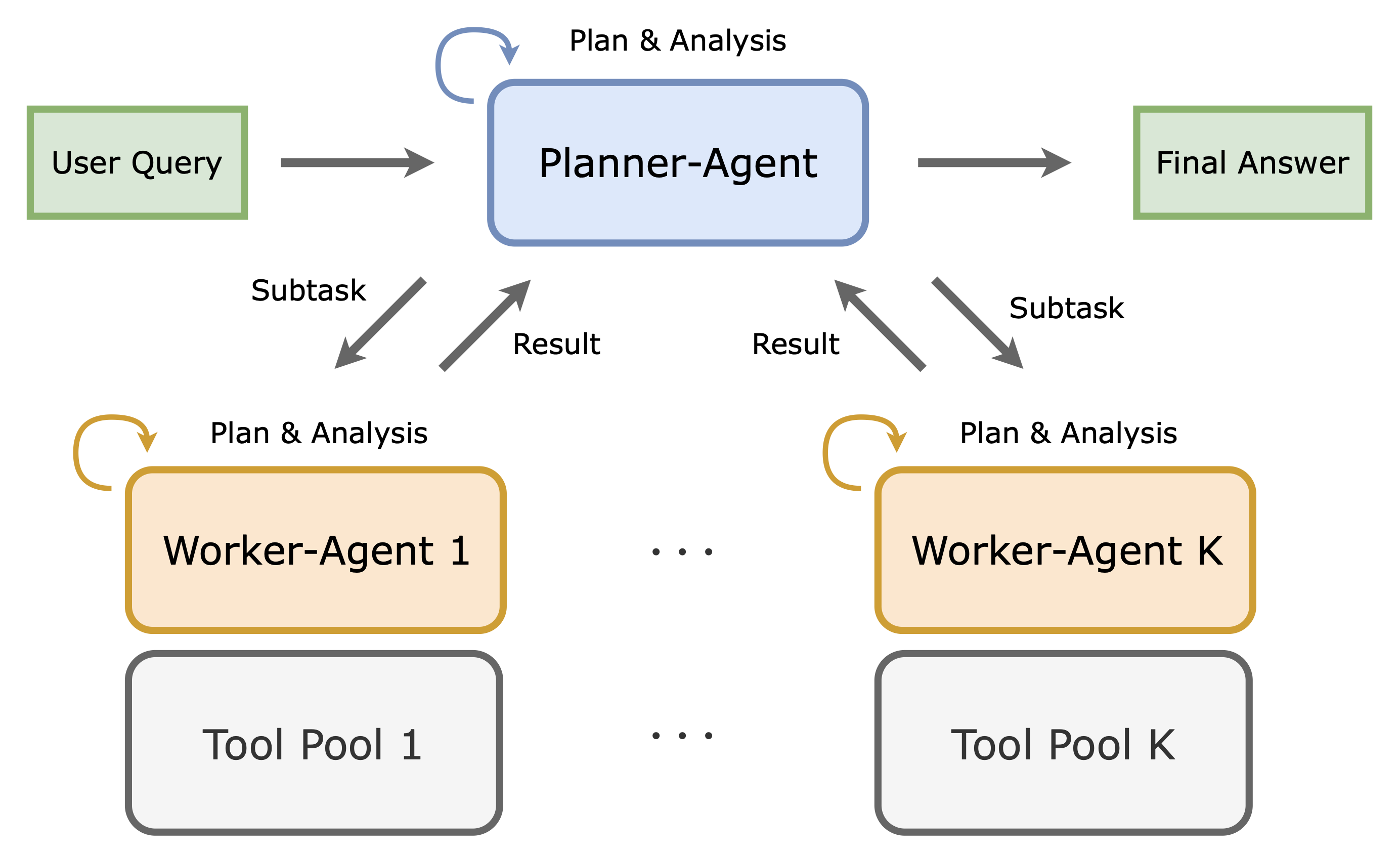}
    \caption{Multi-agent framework. At each step, the planner-agent creates and assigns new subtasks to worker-agents; the planner-agent generates successive subtasks or final answers based on the worker-agents’ responses.}
    \label{fig:multi_agent_framework}
\end{figure}

A straightforward approach to address the above limitations is to use a \textbf{multi-agent} framework \citep{hu2025owl} consisting of a {planner-agent} coordinated with specialized {worker-agent} browsing components, as shown in Figure \ref{fig:multi_agent_framework}.
In the multi-agent framework, the planner-agent orchestrates high-level planning and decision-making while delegating specific browsing tasks to worker-agents, effectively containing noisy search responses within the worker agent’s local context. This allows the planner-agent and worker-agents to maintain manageable context lengths while enabling extended interactions through multiple rounds of coordinated communication and task delegation.

While multi-agent systems offer promising solutions to context and noise management challenges, they introduce new complexities, particularly when each agent operates on separate models. Training such architectures poses significant infrastructure challenges due to uneven workloads across agents, requires substantially more token context, and leads to higher parameter consumption compared to single-agent alternatives.

In this paper, we explore \modelname (\textbf{MATPO}), an algorithm specifically designed for deep research applications, enabling multiple agent roles (\textit{i.e.,} planner- and worker-agents) to coexist within a single model instance. This approach leverages different agent roles activated through distinct system prompts while maintaining the ability to build upon existing reinforcement learning (RL) training frameworks (\textit{e.g.,} veRL \footnote{\url{https://github.com/volcengine/verl}}), preserving the benefits of specialized training while achieving infra efficiency.
We try to address several core research questions in multi-agent RL and system design: 
1. How to perform multi-agent RL training effectively using a single model?
2. How should reward assignment be handled when worker-agents operate without explicit reward signals?
3. Can a single model be used to perform multiple roles, serving as both the planner-agent and worker-agent?

\paragraph{Contributions.}
1. We present \textbf{MATPO}, a principled approach to multi-agent with an end-to-end \textbf{multi-agent-in-one-model} RL training framework;
2. We provide theoretical analysis and a concrete implementation of MATPO;
3. We provide comprehensive experiments to demonstrate that \mn achieves better performance compared to single-agent baselines, accompanied by insights and findings that advance our understanding of multi-agent learning dynamics;
4. We offer practical recommendations for the implementation and training of such systems;
5. We identify meaningful research directions for future exploration in multi-agent RL training.

\section{Related Work}

\subsection{Tool-Integrated Agent Frameworks}
TIP has emerged as a crucial paradigm for enabling LLMs to tackle complex and knowledge-intensive tasks through iterative reasoning combined with external tool use \citep{ve, cok, xu2025comprehensivesurveydeepresearch,openai2025deepresearch}. Building on this advancement, a variety of TIP agent frameworks have been proposed. Early TIP agent frameworks generally follow a single-agent architecture, in which a primary LLM iteratively plans, autonomously invokes tools, such as search APIs or code execution environments, and integrates the tool-responses to refine its reasoning. Representative approaches include function-calling-augmented LLMs \cite{yang2025qwen3,nguyen2025sfrdeepresearcheffectivereinforcementlearning}, ReAct-style agents \citep{react, crplanner,li2025websailornavigatingsuperhumanreasoning,tao2025webshaperagenticallydatasynthesizing}, and agents employing more structured and sophisticated workflows \citep{kimiteam2025kimik2openagentic}.

Despite its simplicity, the single-agent TIP framework faces several fundamental challenges: First, the LLM’s limited context window is quickly saturated by lengthy tool responses and extended multi-turn interaction histories, which hinders scalability to deeper reasoning chains \citep{zhang2025lightthinkerthinkingstepbystepcompression};
Second, tool responses are often noisy or unstructured, and their distribution deviates significantly from that of the LLM’s generation distribution, which can disrupt the LLM's reasoning process and induce cascading reasoning errors \citep{zhou2024languagemodelsperformrobust}.

To mitigate these issues, recent studies have explored multi-agent frameworks \citep{hu2025owl, 2025mirothinker}, where distinct planner- and worker-agents collaborate: the planner performs high-level task decomposition and delegates subtasks to workers, whose responses are then aggregated to produce a final answer.
This decomposition helps contain noisy tool outputs within the worker's local context, allowing the planner to maintain a concise and focused reasoning state across turns.
However, existing efforts only focus on designing sophisticated multi-agent frameworks at inference time via prompt engineering, without providing training methodologies for multi-agent tool-integrated planning. 
\citet{spiral} introduces a framework for training multi-turn multi-agent zero-sum games. However, it is not tailored to the challenges of tool-integrated planning.

\subsection{Tool-integrated Agentic Reinforcement Learning}
Reinforcement learning with verifiable rewards (RLVR) methods have proven effective in training LLMs to improve single-agent TIP performance \citep{deepseekmath,searchr1, 2025mirorl, sfrdeepresearch}.
Beyond standard RLVR, a variety of trajectory filtering techniques have been explored in tasks including math problem solving with code \citep{li2025torlscalingtoolintegratedrl,xue2025simpletirendtoendreinforcementlearning,feng2025retoolreinforcementlearningstrategic} and open-ended
GUI tasks \citep{arpo}. Another line of work starts with supervised fine-tuning (SFT) or direct preference optimization (DPO) \citep{rafailov2024directpreferenceoptimizationlanguage} on cold-start rollout trajectories, and then applies RLVR with carefully designed rewards and rollout strategies, typically within a well-structured TIP agentic workflow \citep{li2025websailornavigatingsuperhumanreasoning,tao2025webshaperagenticallydatasynthesizing,wei2025webagentr1trainingwebagents,Agent-R1,li2025webthinkerempoweringlargereasoning,2025mirorl}.
While these methods have demonstrated notable gains in single-agent settings, principled extensions of RLVR to multi-agent frameworks remain largely underexplored. This highlights the need for training paradigms that efficiently coordinate multiple agent roles, support principled credit assignment, and remain compatible with existing RL infrastructures.

\section{Problem Setup}

\subsection{Single-Agent Multi-Turn Reinforcement Learning}
\begin{figure}[h]
    \centering
    \includegraphics[width=\linewidth]{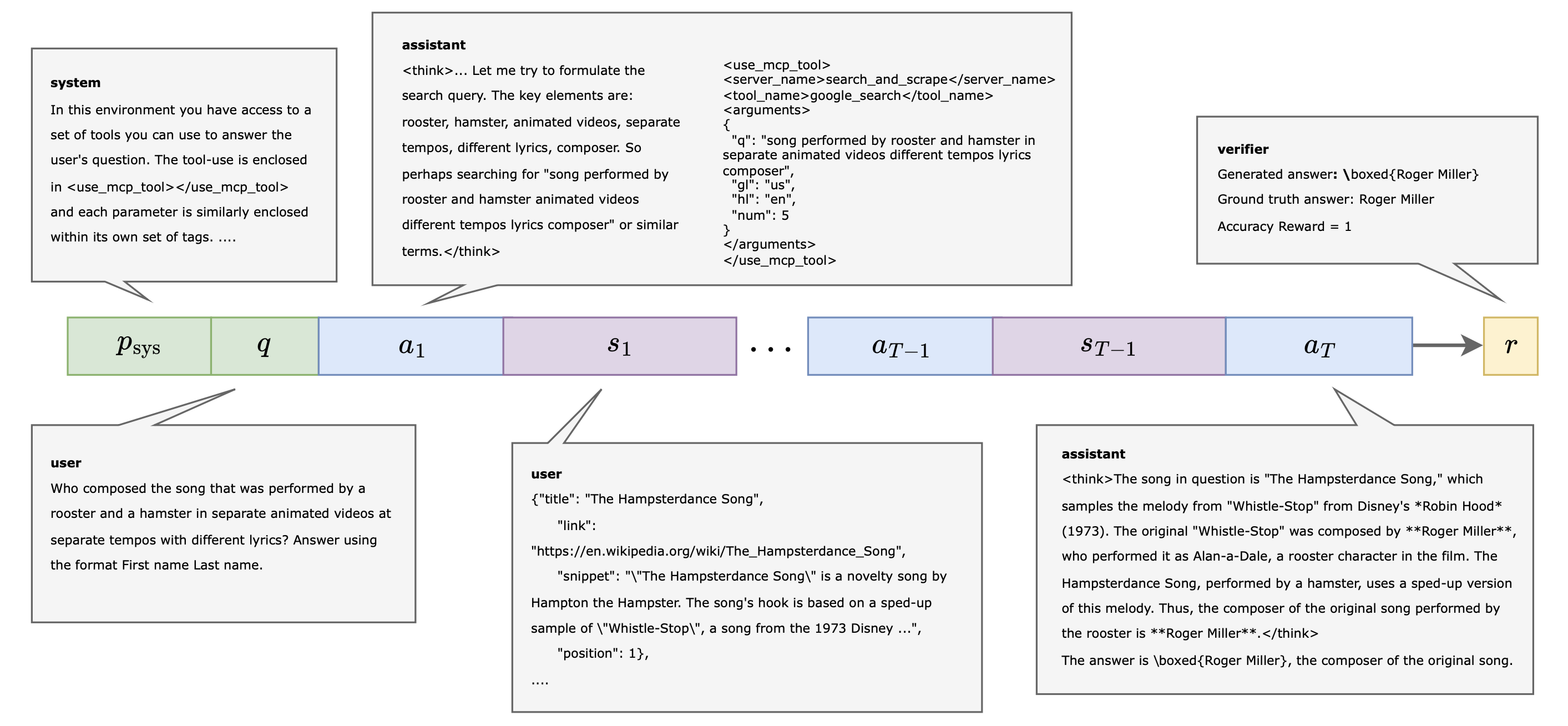}
    \caption{Visualization of a single-agent multi-turn TIP rollout. The LLM solves a query through iterative planning and tool-use. At each step, it plans a tool call, executes it with the parsed parameters, and uses the tool response to decide the next move, continuing until it is confident enough to produce a final answer.}
    \label{fig:single_agent}
\end{figure}

We begin with a brief recap of single-agent multi-turn RL before extending the formulation to the multi-agent setting. Let $\pi_{\theta}(\cdot | \cdot)$ be an LLM  parameterize by $\theta$. For each query $q$ sampled from an underlying distribution $\mathcal{D}$, an LLM agent aims to generate the correct answer to $q$ via a multi-turn tool-integrated planning (TIP) process, as visualized in Figure \ref{fig:single_agent}.

Recent works \citep{toolstar, toolrl} have shown that reinforcement learning with verifiable rewards (RLVR) is a promising approach for enhancing LLMs’ ability to perform the multi-turn TIP process. Given a reward function $r(\cdot)$ that assigns $1$ to correct answers and $0$ to incorrect ones, the objective of single-agent multi-turn RL is
\begin{small}
\begin{align*}
    & \min_{\theta} J(\pi_{\theta})
    \triangleq
    \mathbb{E}_{q\sim \mathcal{D},\tau\sim \pi_{\theta}}
    [r(\tau)],
    \
    \tau \triangleq [a_1,s_1,..,a_T],
    \\
    & a_t \sim
    \pi_{\theta}(\cdot|[p_{\mathrm{sys}},q,a_1,s_1,...,s_{t-1}]),
    \
    s_t \sim\mathrm{Tool}(a_t).
\end{align*}
\end{small}
Specifically, $p_{\mathrm{sys}}$ is the system prompt defining the agent role and tool schema, $a_t$ is the LLM-generated action at turn $t$ including planning and tool-call blocks, $\mathrm{Tool}(\cdot|a_t)$ is the invoked tool conditioned on $a_t$, $s_t$ is its response, and $\tau$ denotes the complete TIP rollout trajectory.

\subsection{Single-Agent Group Relative Policy Optimization}
Among various RL algorithms, GRPO \citep{deepseekmath} has proven to be one of the most effective and efficient methods to minimize $J(\pi_{\theta})$. To adapt GRPO to the single-agent multi-turn TIP setting, note that each rollout includes both the LLM-generated tokens $a_1,...,a_T$ (the blue blocks in Figure \ref{fig:single_agent}) and tool-API response tokens $s_1,...,s_T$ (the purple blocks in  Figure \ref{fig:single_agent}). As the tool-response tokens are not generated by $\pi_\theta$, they do not contribute to the policy gradient for the GRPO objective. Therefore, the single-agent GRPO objective masks out all tool-response tokens as follows:
\begin{small}
\begin{align*}
& J_{\mathrm{single}}(\pi_{\theta})
\triangleq 
\mathbb{E}_{\substack{
q\sim\mathcal{D}\\
\{\tau_i\}_{i=1}^G \sim \pi_{\theta_{\mathrm{old}}}
}}
\left[
\frac{1}{G}\sum_{i=1}^G \frac{1}{\sum_{t=1}^{T_i} |a_t^i| }
\sum_{t=1}^{T_i} R^{\mathrm{clip}}_i
\right]
\\
& R^{\mathrm{clip}}_i
\triangleq
\min(R_{i,t}(\theta) \hat{A}_{i,t}, \mathrm{clip}(R_{i,t}(\theta),1-\varepsilon,1+\varepsilon)\hat{A}_{i,t}),
\\
& R_{i,t}(\theta)\triangleq  \frac{\pi_{\theta}(a_{t}^i|[p_{\mathrm{sys}},q,a_1^i,s_1^i,...,s_{t-1}^i])}{\pi_{\theta_{\mathrm{old}}}(a_{t}^i|[p_{\mathrm{sys}},q,a_1^i,s_1^i,...,s_{t-1}^i])},
\\
&\hat{A}_{i,t}\triangleq (r(\tau_i) - \mathrm{mean}(\{r(\tau_i)\}_{i=1}^G))/\mathrm{std}(\{r(\tau_i)\}_{i=1}^G),
\end{align*}
\end{small}

where $\pi_{\theta_{\mathrm{old}}}$ denotes a periodically updated snapshot of the target LLM $\pi_{\theta}$, and $\pi_{\mathrm{ref}}$ is a fixed reference model (e.g., the checkpoint from which RL training begins). $G$ denotes the group size of rollouts associated with each query $q$. Each rollout is represented as $\tau_i \triangleq [a_1^i, s_1^i, \ldots, a_{T_i}^i]$, comprising $T_i$ turns, with $\sum_{t=1}^{T_i} |a_t^i|$ indicating the total number of LLM-generated tokens. $R_{i,t}(\theta)$ represents the likelihood ratio of action $a_t^i$ between $\pi_{\theta}$ and $\pi_{\theta_{\mathrm{old}}}$, $\hat{A}_{i,t}$ is the group-relative normalized reward, and $\mathrm{clip}(\cdot, 1-\varepsilon, 1+\varepsilon)$ is the clipping function restricting values to $[1-\varepsilon, 1+\varepsilon]$.


\subsection{Multi-Agent Multi-Turn Reinforcement Learning}
\begin{figure}[h]
    \centering
\includegraphics[width=\linewidth]{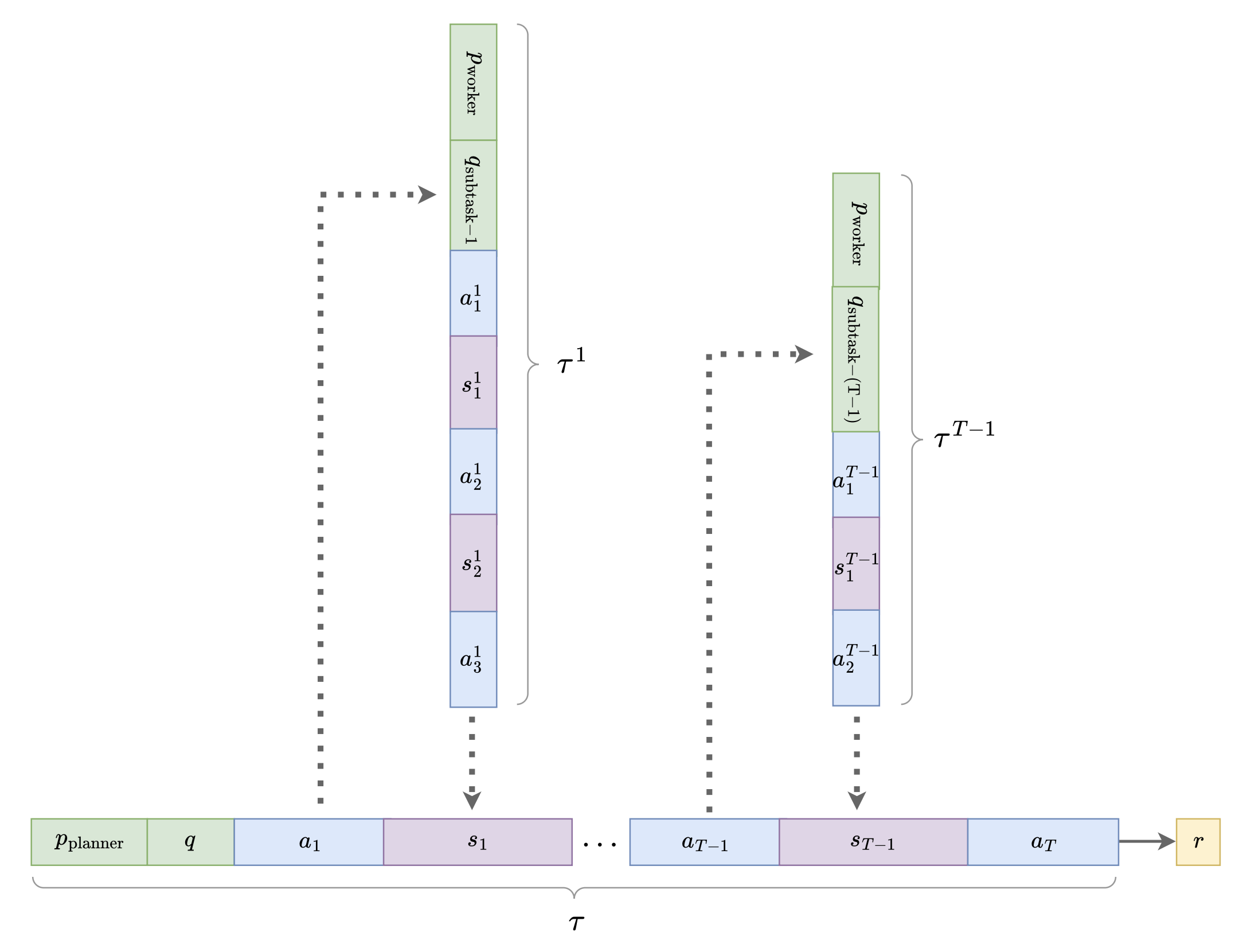}
    \caption{Visualization of a multi-agent multi-turn TIP rollout. At each step, the planner agent generates and assigns a subtask to the worker agent, which completes it via multi-turn TIP and returns the result. The planner agent then decides whether to generate a new subtask or produce the final answer based on this response.}
    \label{fig:multi_agent_rl_rollout}
\end{figure}

As mentioned in the introduction, multi-agent multi-turn TIP frameworks are designed to overcome the context length bottleneck and noisy tool-token issues present in single-agent multi-turn TIP. 
For clarity and without loss of generality, this paper considers a multi-agent framework with one planner-agent and one worker-agent. A multi-agent multi-turn TIP rollout is visualized in Figure \ref{fig:multi_agent_rl_rollout}. Specifically define $q$ denotes the user query, and $\tau$ represents the entire multi-turn TIP rollout for handling it. $p_{\mathrm{planner}}$ is the system prompt specifying the role of the planner agent. At each turn $t$, the planner generates an action $a_t$ containing a thinking block and either a subtask or the final answer, and receives a response $s_t$ parsed from the worker agent’s output. The planner proceeds for $T$ turns in total. Each subtask query $q_{\mathrm{subtask-}t}$ parsed from $a_t$ is handled by a worker-agent rollout $\tau^t$, guided by the system prompt $p_{\mathrm{worker}}$. Within $\tau^t$, the worker produces actions $a_i^t$ (each including a thinking block and either a tool call or a final sub-answer) and receives tool responses $s_i^t$. Finally, $r$ denotes the accuracy reward for the final planner answer $a_T$.


As shown in Figure \ref{fig:multi_agent_rl_rollout}, each multi-agent TIP rollout consists of $T$ single-agent TIP rollouts: one from the planner agent and $(T-1)$ from worker agents handling their respective subtasks. Specifically, a multi-agent TIP rollout is 
\begin{small}
\begin{align*}
\tau\triangleq &[a_1,\tau^1,s_1,...,a_{T-1},\tau^{T-1},s_{T-1},a_T]\sim(\pi_{\theta},\mathrm{Tool}),
\\
\tau^t
\triangleq &
[a_{1}^t,s_{1}^t,...,s_{T_t-1}^t,a_{T_t}^t],
\ 
s_t\sim \mathrm{Parse}(a_{T_t}^t), s_{i}^t\sim \mathrm{Tool}(a_{i}^t).
\end{align*}
\end{small}
where $\mathrm{Parse}(a_{T_t}^t)$ is the worker-agent’s response to the $t$-th subtask parsed from the final content in the worker-agent rollout, and $\mathrm{Tool}(a_i^t)$ is the tool-response based on the parameters parsed from action $a_i^t$ from the worker-agent.

Given a reward function $r(\cdot)$ that assigns $1$ to correct answers and $0$ to incorrect ones, the objective of multi-agent multi-turn RL can be formalized as:
\begin{small}
\begin{align*}
& \min_{\theta} J_{\mathrm{multi}}(\pi_{\theta})
\triangleq
\mathbb{E}_{q\sim \mathcal{D},\tau\sim (\pi_{\theta},\mathrm{Tool})}
[r(\tau)],
\\
& a_t \sim
\pi_{\theta}(\cdot\ |[p_{\mathrm{planner}},q,a_1,s_1,...,s_{t-1}]),
\\
& a_{j}^t \sim
\pi_{\theta}(\cdot\ |[p_{\mathrm{worker}},q_{\mathrm{subtask}-t},a_{1}^t,s_{1}^t,...,s_{j-1}^t]),
\\
& s_{t} \sim \mathrm{Parse}(a_{T_t}^t),
\
q_{\mathrm{subtask}-t}\sim \mathrm{Parse}(a_t),
\
s_{j}^t \sim \mathrm{Tool}(a_{j}^t).
\end{align*}
\end{small}

Notice that in $J_{\mathrm{multi}}(\pi_{\theta})$, a single LLM $\pi_{\theta}$ is deployed to serve as both the planner-agent and the worker-agent, distinguished only by different system prompts $p_{\mathrm{planner}}$ and $p_{\mathrm{worker}}$. In this paper, we refer to this deployment configuration as \textbf{multi-agent-in-one-model}.

An alternative configuration is to deploy separate models for the planner-agent and worker-agents, which we refer to as multi-agent-multi-model. The multi-agent multi-turn RL objective can be directly generalized to this configuration. Let the planner-agent be parameterize by $\pi_{\theta}$ and $K$ worker-agents parameterize by $\pi_{\phi_1},...,\pi_{\phi_K}$. The resulting multi-agent-multi-model objective is
\begin{small}
\begin{align*}
& J_{\mathrm{multi}}(\pi_{\theta},\textcolor{blue}{\{\pi_{\phi_k}\}_{k\in[K]}})
\triangleq
\mathbb{E}_{q\sim \mathcal{D},\tau\sim (\pi_{\theta},\textcolor{blue}{\{\pi_{\phi_k}\}_{k\in[K]}},\mathrm{Tool})}
[r(\tau)],
\\
& a_t \sim
\pi_{\theta}(\cdot\ |[p_{\mathrm{planner}},q,a_1,s_1,...,s_{t-1}]),
\\
& a_{j}^t \sim
\textcolor{blue}{\pi_{\phi_k}}(\cdot\ |[p_{\mathrm{worker}},q_{\mathrm{subtask}-t},a_{1}^t,s_{1}^t,...,s_{j-1}^t]),
\ 
k\in [K],
\\
& s_{t} \sim \mathrm{Parse}(a_{T_t}^t),
\
(q_{\mathrm{subtask}-t},\textcolor{blue}{k})\sim \mathrm{Parse}(a_t),
\
s_{j}^t \sim \mathrm{Tool}(a_{j}^t).
\end{align*}
\end{small}

In this paper, we focus on exploring RL training under the multi-agent-in-one-model setting, as it offers several advantages over the multi-agent-multi-model setting: 
1) the multi-agent-multi-model setting requires $(K+1)$ LLM rollout engines and additional RL infrastructure optimization. In contrast, the multi-agent-in-one-model framework uses only ONE single LLM rollout engine and remains compatible with off-the-shelf RL frameworks;
2) We are interested in whether RL training can benefit the model when it is exposed to experience from multiple agent roles.

\section{Methodology}
\subsection{Multi-Agent Tool-Integrated Policy Optimization}
A key challenge in extending single-agent GRPO to the multi-agent setting is \textbf{credit assignment}: how should the planner-agent rollout $\tau^0$ and the worker-agent rollouts $\tau^t$ share responsibility for the final accuracy of the full multi-turn TIP rollout $\tau$? The planner-agent's final answer is directly verifiable, whereas worker-agent rollouts address unverifiable subtasks, making it essential to assess their contribution to the planner's final answer.

In this section, we derive the GRPO counterpart in the multi-agent-in-one-model setting to optimize $J_{\mathrm{multi}}(\pi_{\theta})$. Notice that the policy gradient $\nabla_{\theta} J_{\mathrm{multi}}(\pi_{\theta})$ equals to
\begin{small}
\begin{align*}
& \nabla_{\theta} J_{\mathrm{multi}}(\pi_{\theta})
=
\nabla_{\theta}
\mathbb{E}_{q\sim \mathcal{D},\tau\sim (\pi_{\theta},\mathrm{Tool})}
[r(\tau)]
\\
= &
\mathbb{E}_{q\sim \mathcal{D},\tau\sim (\pi_{\theta},\mathrm{Tool})}
[r(\tau) \nabla_{\theta} \log \mathbb{P}_{\theta}(\tau)],
\end{align*}
\end{small}
where $r(\tau)$ denotes the accuracy reward associate to the full multi-agent multi-turn TIP rollout $\tau$, $\mathbb{P}_{\theta}(\tau)$ denotes the probability of generating $\tau$ using LLM $\pi_{\theta}$.This implies
\begin{small}
\begin{align*}
& \mathbb{P}_{\theta}(\tau)
\triangleq
\mathbb{P}_{\theta}
(
[p_{\mathrm{planner}},q,a_1,\tau^1,s_1,...,\tau^{T-1},s_{T-1},a_T]
)
\\
= &
\pi_{\theta}(a_1|[p_{\mathrm{planner}},q])
\mathbb{P}_{\theta}(\tau^1|a_1)
\cdots
\mathbb{P}_{\theta}(\tau^{T-1}|a_{T-1})
\\
& \cdot\pi_{\theta}(a_T|[p_{\mathrm{planner}},q,a_1,...,s_{T-1}]),
\\[1mm]
& \mathbb{P}_{\theta}(\tau^t|a_t)
\triangleq 
\mathbb{P}_{\theta}
(
[p_{\mathrm{worker}},q_{\mathrm{subtask}-t},a_1^t,s_1^t,...,s_{T_t-1}^t,a_{T_t}^t]
)
\\
= &
\pi_{\theta}(a_1|[p_{\mathrm{worker}},q_{\mathrm{subtask}-t}])
\mathbb{P}_{\mathrm{Tool}}(s_1|a_1)
\\
& \cdot
\pi_{\theta}(a_2|[p_{\mathrm{worker}},q_{\mathrm{subtask}-t},a_1,s_1]) \cdots
\mathbb{P}_{\mathrm{Tool}}(s_{T-1}|a_{T-1})
\\
& \cdot \pi_{\theta}(a_T|[p_{\mathrm{worker}},q_{\mathrm{subtask}-t},q,a_1,...,s_{T-1}]).
\end{align*}
\end{small}
As the tool-responses are not generated by the LLM $\pi_{\theta}$, it holds that $\nabla_{\theta} \mathbb{P}_{\mathrm{Tool}}(s_t|a_t) = 0$, and
\[
\resizebox{\columnwidth}{!}{$
\begin{aligned}
& \nabla_{\theta} \log \mathbb{P}_{\theta}(\tau)
= \nabla_{\theta} \Big(
\log \pi_{\theta}(a_1|[p_{\mathrm{planner}},q])
+ \log \mathbb{P}_{\theta}(\tau^1|a_1)
+ \cdots
\\
& + \log \mathbb{P}_{\theta}(\tau^{T-1}|a_{T-1})
+ \log \pi_{\theta}(a_T|[p_{\mathrm{planner}},q,a_1,...,s_{T-1}])
\Big)
\\[1mm]
= & \sum_{t=1}^T \nabla_{\theta} \log \pi_{\theta}(a_t|[p_{\mathrm{planner}},q,a_1,s_1,..,s_{t-1}])
+ \sum_{t=1}^{T-1} \nabla_{\theta} \log \mathbb{P}_{\theta}(\tau^t|a_t)
\\[1mm]
= & \sum_{t=1}^T 
\frac{\nabla_{\theta} \pi_{\theta}(a_t|[p_{\mathrm{planner}},q,a_1,s_1,..,s_{t-1}])}
{\pi_{\theta}(a_t|[p_{\mathrm{planner}},q,a_1,s_1,..,s_{t-1}])}
\\
& + \sum_{t=1}^{T-1} \sum_{j=1}^{T_t} 
\frac{\nabla_{\theta} \pi_{\theta}(a_j^t|[p_{\mathrm{worker}},q_{\mathrm{subtask}-t},a_1^t,s_1^t,..,s_{j-1}^t])}
{\pi_{\theta}(a_j^t|[p_{\mathrm{worker}},q_{\mathrm{subtask}-t},a_1^t,s_1^t,..,s_{j-1}^t])}.
\end{aligned}
$}
\]
where $\tau^0\triangleq [p_{\mathrm{planner}},q,a_1,s_1,..,s_{T-1},a_T]$ denotes the rollout trajectory of the planner-agent and $\tau^t$ is exactly the $t$-th rollout trajectory of the worker-agent associated to the $t$-th subtask.

Following the standard derivation of vanilla GRPO, we can derive the \mn objective as:
\begin{small}
\begin{align*}
& J_{\mathrm{MATPO}}(\pi_{\theta})
\triangleq
\mathbb{E}_{\substack{q\sim \mathcal{D}\\\{\tau_i\}\sim (\pi_{\theta_{\mathrm{old}}},\mathrm{Tool})}}
\left[
\frac{1}{G}
\sum_{i=1}^G
\frac{1}{\sum_{t=0}^{T_i} |\tau_i^t|}
\sum_{t=0}^{T_i}
R^{\mathrm{clip}}_i
\right]
\\
& R^{\mathrm{clip}}_i\triangleq
\min(
R_{i,t}\hat{A}_{i,t},
\mathrm{clip}(R_{i,t},1-\varepsilon,1+\varepsilon)
\hat{A}_{i,t}
)
\\
& \hat{A}_{i,t} \triangleq (r(\tau_i) - \mathrm{mean}(\{r(\tau_i)\}_{i=1}^G)) / \mathrm{std}(\{r(\tau_i)\}_{i=1}^G)
\end{align*}
\end{small}
where $\tau_i$ denotes the full multi-agent TIP rollout for the $i$-th query $q$, containing $T_i$ subtasks; we denote $\tau_i^0$ as the planner-agent rollout and $\tau_i^t$ ($t>0$) as the $t$-th worker-agent rollout within $\tau_i$; $\hat{A}_{i,t}$ denotes the group-relative normalized reward among $G$ full rollouts. Specifically, $R_{i,t}$ defines the log-likelihood ratio between $\pi_{\theta_{\mathrm{old}}}$ and $\pi_\theta$ of $\tau_i$, defined as
\[
\resizebox{\columnwidth}{!}{$
\begin{aligned}
R_{i,t} \triangleq
\begin{cases}
\sum_{j=1}^{T_i} 
\dfrac{
\pi_{\theta_{\mathrm{old}}}(a_j^t|[p_{\mathrm{planner}},q,a_1,s_1,..,s_{j-1}])
}{
\pi_{\theta}(a_j^t|[p_{\mathrm{planner}},q,a_1,s_1,..,s_{j-1}])
}, & t=0,\\
\sum_{j=1}^{T_{i,t}} 
\dfrac{
\pi_{\theta_{\mathrm{old}}}(a_j^t|[p_{\mathrm{worker}},q_{\mathrm{subtask}-t},a_1^t,s_1^t,..,s_{j-1}^t])
}{
\pi_{\theta}(a_j^t|[p_{\mathrm{worker}},q_{\mathrm{subtask}-t},a_1^t,s_1^t,..,s_{j-1}^t])
}, & t>0,
\end{cases}
\end{aligned}
$}
\]
where $T_{i,t}$ is the tool-calls count in the $t$-th subtask of $\tau_i$.

We summarize the key distinctions between single-agent GRPO and \mn as follows: unlike GRPO, which performs a single worker-agent rollout per update, \mn executes one planner-agent rollout followed by $T$ worker-agent rollouts. Moreover, while GRPO normalizes rewards across $G$ worker rollouts for credit assignment, \mn normalizes across $G \times (T + 1)$ rollouts to jointly account for planner and worker contributions.



\subsection{Implementation}
\begin{figure}[h]
    \centering
    \includegraphics[width=\linewidth]{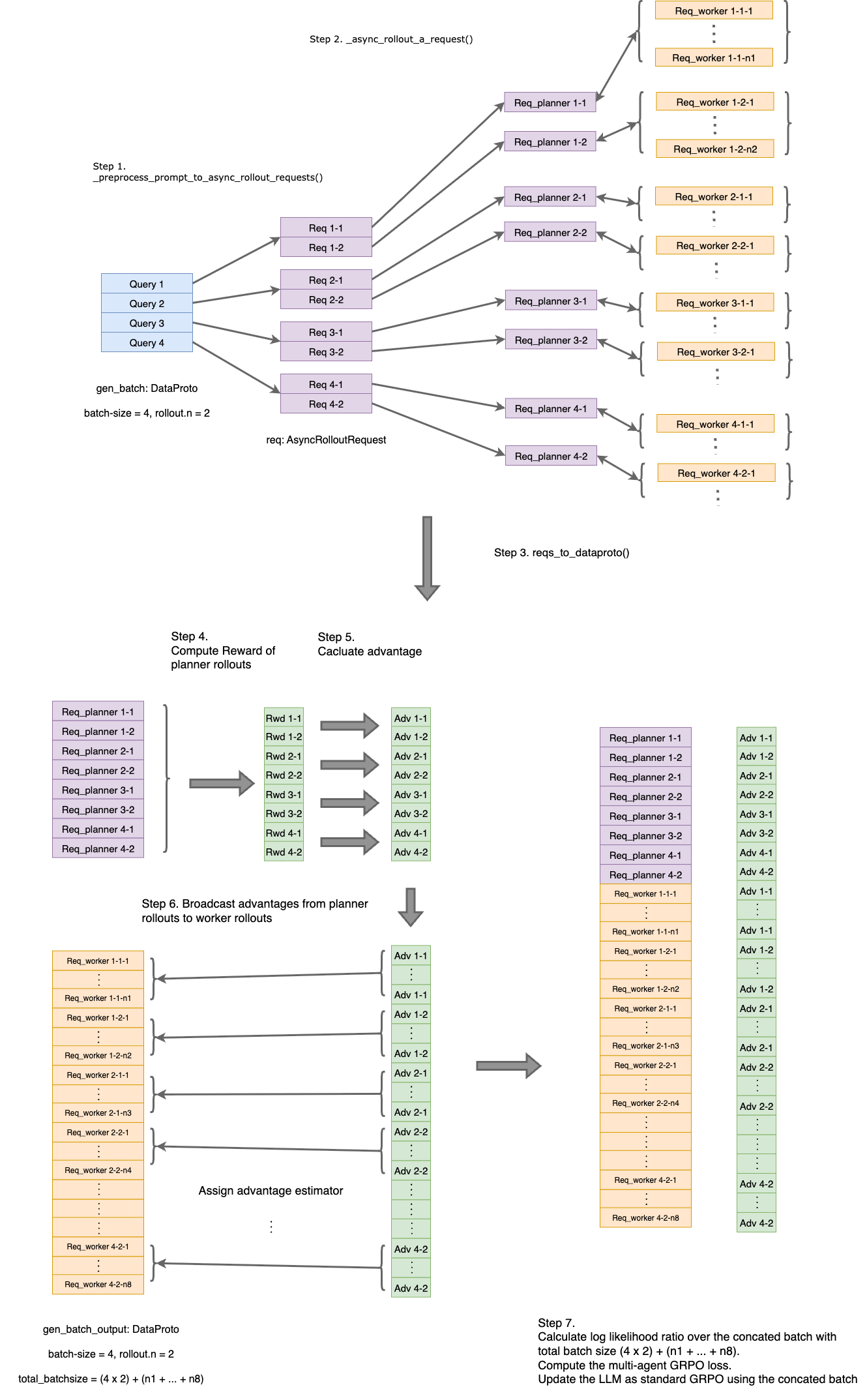}
    \caption{An illustration of the implementation of MATPO.}
    \label{fig:multi_agent_grpo_imp}
\end{figure}

Figure \ref{fig:multi_agent_grpo_imp} provides an illustrative visualization of the implementation of MATPO, showing how it can be built upon single-agent multi-turn RL frameworks.

For each user query, we first feed \texttt{n.rollout} rollout requests to the rollout engine (\textit{e.g.,} \texttt{vLLM} or \texttt{sglang}).
Next, we modify the original rollout function so that when a worker-agent is invoked, a nested rollout function is launched within the outer one, and these processes execute asynchronously. 
For each query, we generate \texttt{n.rollout} planner-agent rollouts (the purple boxes in Figure \ref{fig:multi_agent_grpo_imp}), with each one associated with a bundle of worker-agent rollouts (the orange boxes enclosed in the braces in Figure \ref{fig:multi_agent_grpo_imp}) generated to tackle the subtasks assigned by their respective planner-agents. 
Then, both the planner-agent and worker-agent rollouts are converted from rollout requests to data batches.
After that, for each planner-agent rollout (the purple boxes), we compute its accuracy reward by verifying whether its final answer block reveals the ground truth answer to the user query. 
Following this, we compute advantages by normalizing this accuracy reward among the group of planner-agent rollouts associated with each user query. 
Subsequently, the computed advantages for a planner-agent rollout are then broadcast to its corresponding worker-agent rollouts.
Finally, we concatenate the planner-agent rollouts and the worker-agent rollouts into an augmented batch (the stack comprising both purple and orange boxes on the right). 
We compute the log likelihood on this augmented batch using $\pi_{\theta}$ and $\pi_{\theta_{\mathrm{old}}}$. With this, we compute the loss, $J_{\mathrm{MATPO}}(\pi_{\theta})$, and mask out the entries of all tokens from agent system prompts, the query, and tool responses. The LLM $\pi_{\theta}$ is then updated using the augmented batch through the standard optimization process.

\section{Experiments}
\subsection{Setups}
In this work, we focus on the deep search scenario, where a planner-agent and a worker-agent comprise a two-agent system, aiming to find the answer of a given user query based on searching and web scraping \footnote{To avoid potential leakage of datasets hosted on HuggingFace, search results from this site are blocked by default, unless noted.
}. Specifically We implement our algorithm on top of veRL \footnote{\url{https://github.com/volcengine/verl}}. The training hyperparameters are provided in the training script released in the GitHub repository. All experiments are conducted with 128 A800 GPUs.
In this section, we introduce the implementation details of our proposed MATPO.

\paragraph{Dataset and Base Model.}
All experiments are conducted on the \texttt{Qwen3-14B-base} model. We train the model with either single-agent GRPO or \mn on a filtered subset of the \texttt{MuSiQue} \citep{musique} dataset, a multi-hop QA dataset. We remove overly difficult queries for which LLMs repeatedly fail to produce valid rollouts. 
Our models are then tested on \texttt{GAIA-text} \citep{gaia} \footnote{\texttt{GAIA-text} is a curated subset of 103 text-only queries drawn from the \texttt{GAIA} dataset \cite{gaia}, a benchmark for general AI assistants.}, \texttt{WebWalkerQA} \citep{webwalker}, and \texttt{FRAMES} \citep{frames}.

\paragraph{Agent System Prompt and Tool-Call Format.}
We use an XML format to parse tool calls from both planner and worker agents. The planner-agent's system prompt specifies the tool schema to call the worker-agent, while the worker-agent's system prompt specifies schema of tool-calls of Google's Serper API for search and scraping. After each tool call, the tool's responses are wrapped as a ``user message'' and appended to the agent's rollout trajectory. To help the worker agent execute the user’s original query from the planner agent, we include a recap of the query in the worker agent’s system prompt, a process we call “user query recapping.” The detailed system prompts and tool schemas of the planner- and worker-agents are in Appendix.

\paragraph{Reward Function.}
In this work, we use LLM-as-a-judge \footnote{We implement the LLM-as-judge based on GPT-4o-mini with instructions shown in Appendix.} to evaluate the accuracy of a model’s answer against the ground-truth answer. The RL reward is set as \lstinline[breaklines=true]{reward = 0.9 * acc + 0.1 * fmt}, where \lstinline[breaklines=true]{acc} is a binary value indicating whether the rollout is correct, \lstinline[breaklines=true]{fmt} measures the average correctness of the tool-calls generated by the model. Specifically, for single-agent RL, we define \lstinline[breaklines=true]{fmt} as the success rate of all tool-call attempts parsed from the LLM’s generated action. For MATPO, we define \lstinline[breaklines=true]{fmt= 0.5 * fmt_p + 0.5 * fmt_w}, where \lstinline[breaklines=true]{fmt_p}  denotes the successful tool-call rate among a planner-agent rollout, and \lstinline[breaklines=true]{fmt_w} denotes the average successful tool-call rate among all associated worker-agent rollouts.

\paragraph{Rollout Summary Mechanism.} 
To encourage the agent to generate answers based on the entire rollout trajectory, we implement a final-summary mechanism. At the end of each rollout, we instruct the model to stop further tool calls and produce an answer based on a summary of the full rollout. We then perform an additional round of summarization and append this final summary to the complete rollout trajectory. \footnote{Rollout summary prompt is detailed in Appendix.}
To avoid exceeding the model’s maximum token length, if a rollout reaches the limit, we remove the latest messages from the trajectory until there is sufficient token budget for the final summary.  Both of the worker-agents in single-agent and multi-agent RL settings are equipped with such summary mechanism.

\begin{figure}[h]
    \centering 
    \begin{subfigure}[b]{0.3\textwidth}
        \centering
        \includegraphics[width=\textwidth]{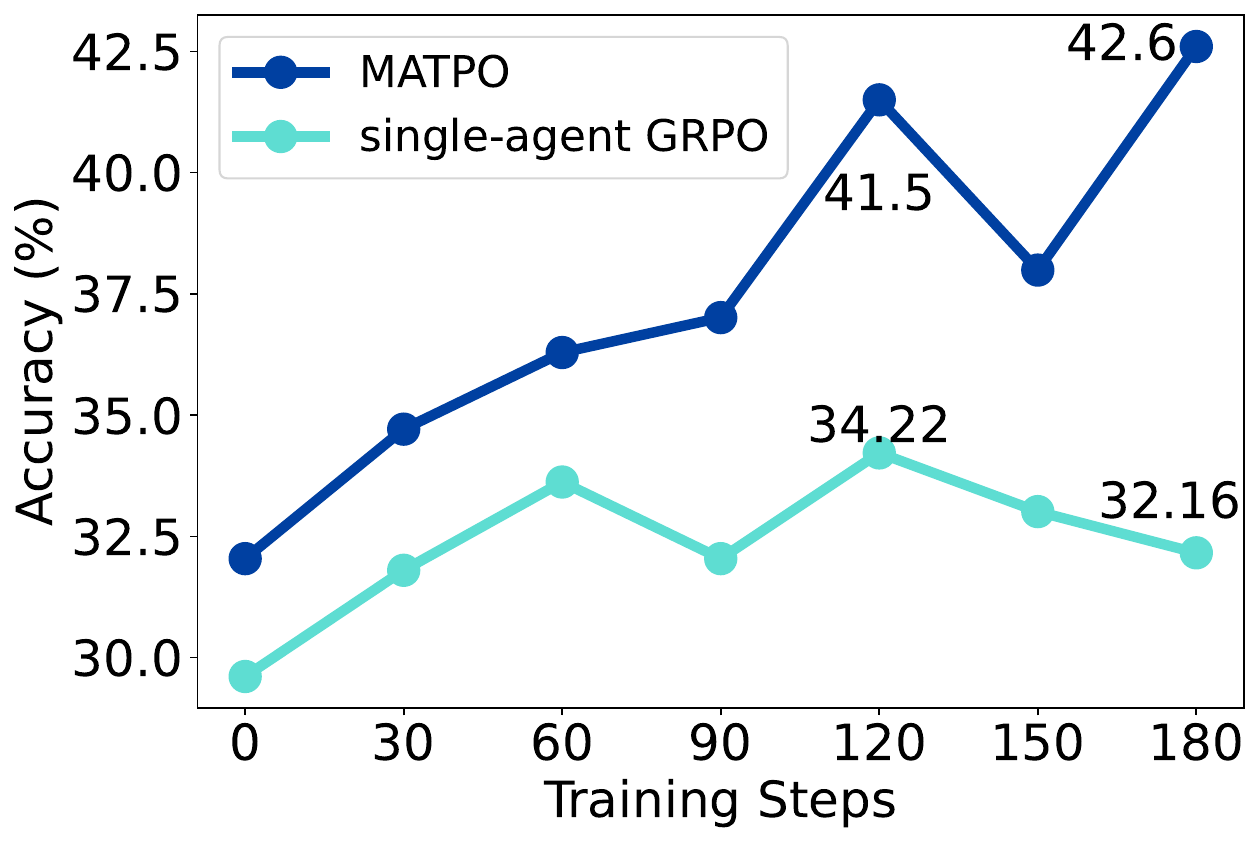}
        \caption{Test accuracy on the \texttt{GAIA-text} dataset \cite{gaia}.} 
        \label{fig:main-exp-gaia} 
    \end{subfigure}
    \\
    \begin{subfigure}[b]{0.3\textwidth}
        \centering
        \includegraphics[width=\textwidth]{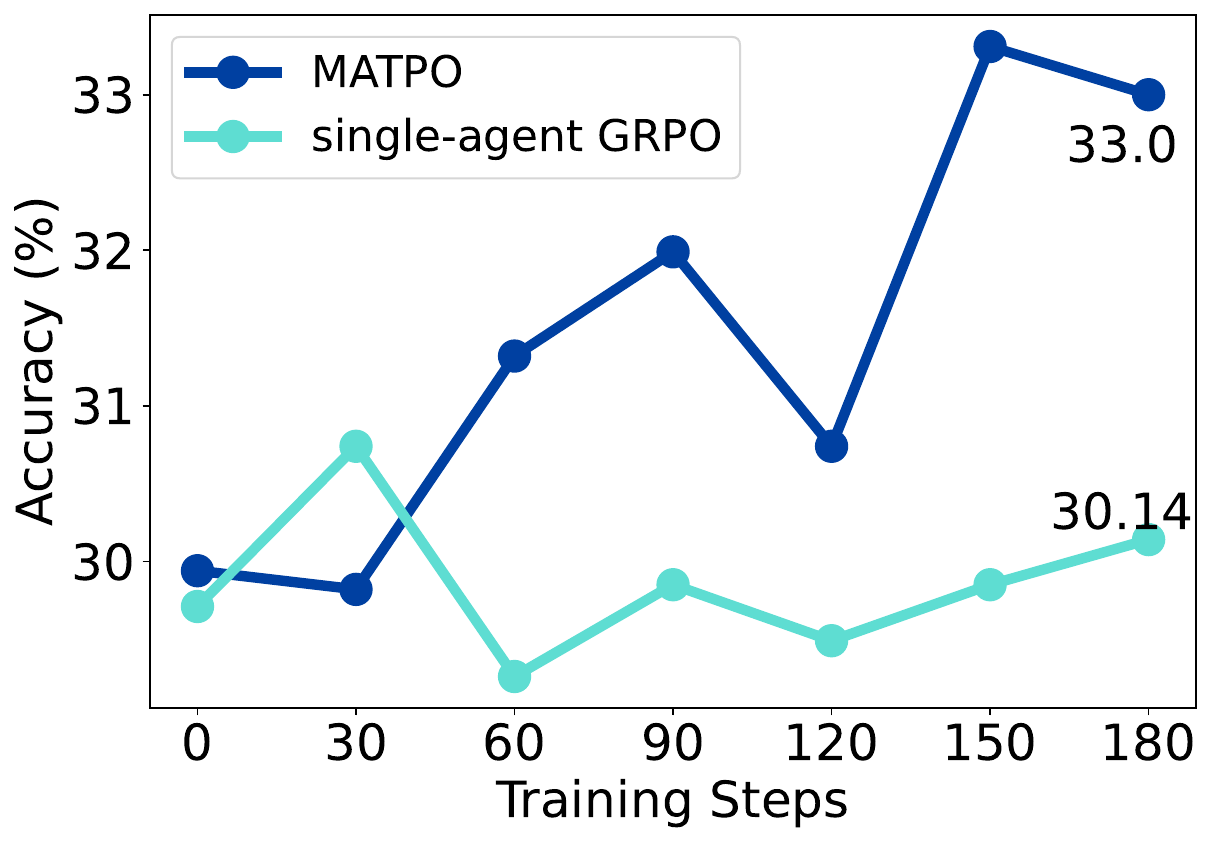}
        \caption{Test accuracy on the \texttt{WebWalkerQA} dataset \cite{webwalker}.} 
        \label{fig:main-exp-webwalker} 
    \end{subfigure}
    \\
    \begin{subfigure}[b]{0.3\textwidth}
        \centering
        \includegraphics[width=\textwidth]{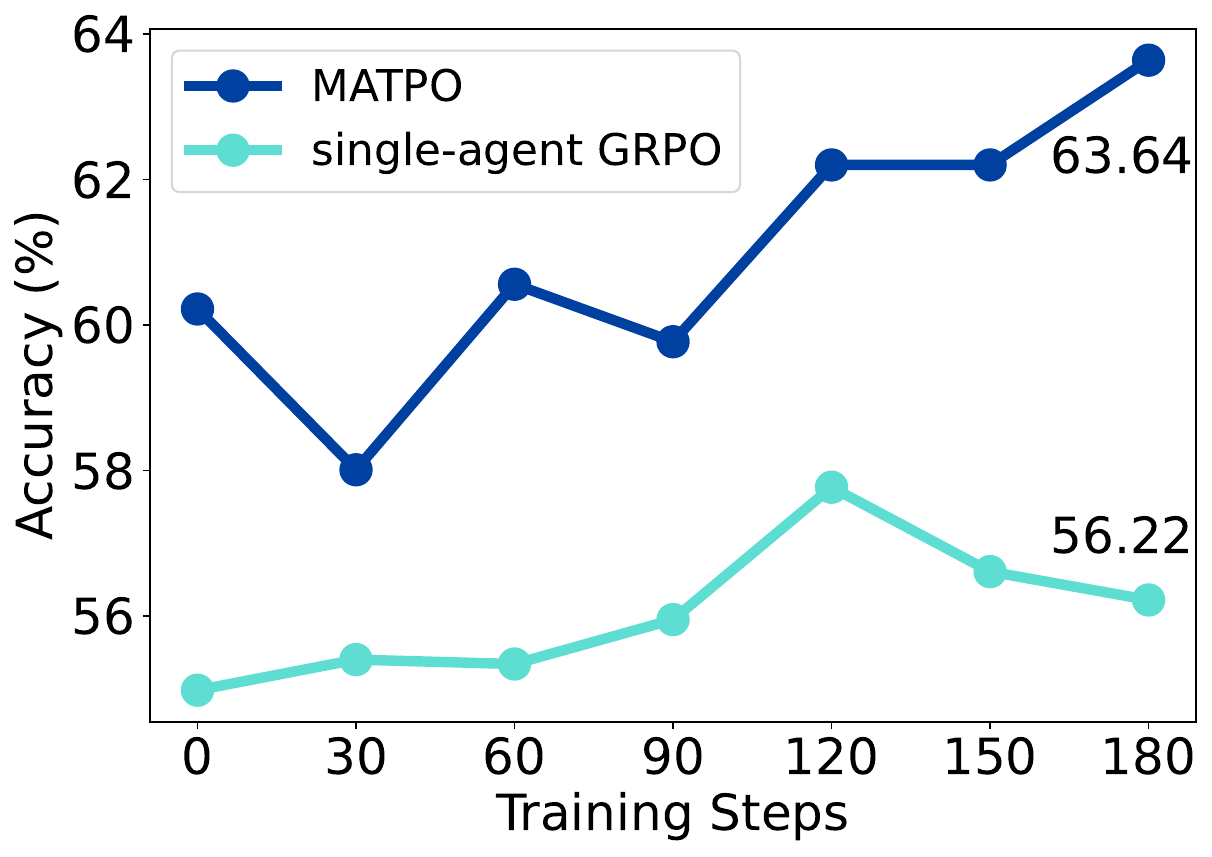}
        \caption{Test accuracy on the \texttt{FRAMES} dataset \cite{frames}.} 
        \label{fig:main-exp-frame} 
    \end{subfigure}
    \caption{Test accuracy on three benchmarks across different training steps. Models are trained on the \texttt{MusiQue} dataset \cite{musique}.}
    \label{fig:main-exp} 
\end{figure}

\subsection{Results}
\paragraph{MATPO consistently outperforms single-agent GRPO.}

Figure \ref{fig:main-exp} presents the testing accuracy on \texttt{GAIA-text}, \texttt{WebWalkerQA}, and \texttt{FRAMES} across different training steps. 
MATPO consistently surpasses the single-agent GRPO baseline, underscoring the effectiveness of our approach. 
Specifically, MATPO achieves $42.60\%$, $33.00\%$, and $63.64\% $on \texttt{GAIA-text}, \texttt{WebWalkerQA}, and \texttt{FRAMES}, respectively, compared to $32.16\%$, $30.14\%$, and $56.22\%$ for single-agent GRPO, leading to an average relative improvement of $18.38\%$.
Moreover, MATPO exhibits more stable gains as training progresses. For instance, while the performance of single-agent GRPO drops after step $120$ on both \texttt{GAIA-text} and \texttt{FRAMES}, MATPO continues to improve.
We attribute this divergence to the vulnerability of single-agent training: agentic RL often suffers catastrophic drops in performance due to unstable environmental feedback (\textit{e.g.,} missing or noisy responses from the Serper API). In contrast, MATPO can invoke additional browinsg subtasks, enabling the agent to perform more robust searches and maintain steady progress.


\subsection{Ablation Studies and Practical Take-Aways}
We conduct ablation studies on the key components of MATPO and summarize implementation techniques that enhance its stability and performance.
\begin{figure}[h]
    \centering 
    \begin{subfigure}[b]{0.48\textwidth}
        \centering
    \includegraphics[width=\textwidth]{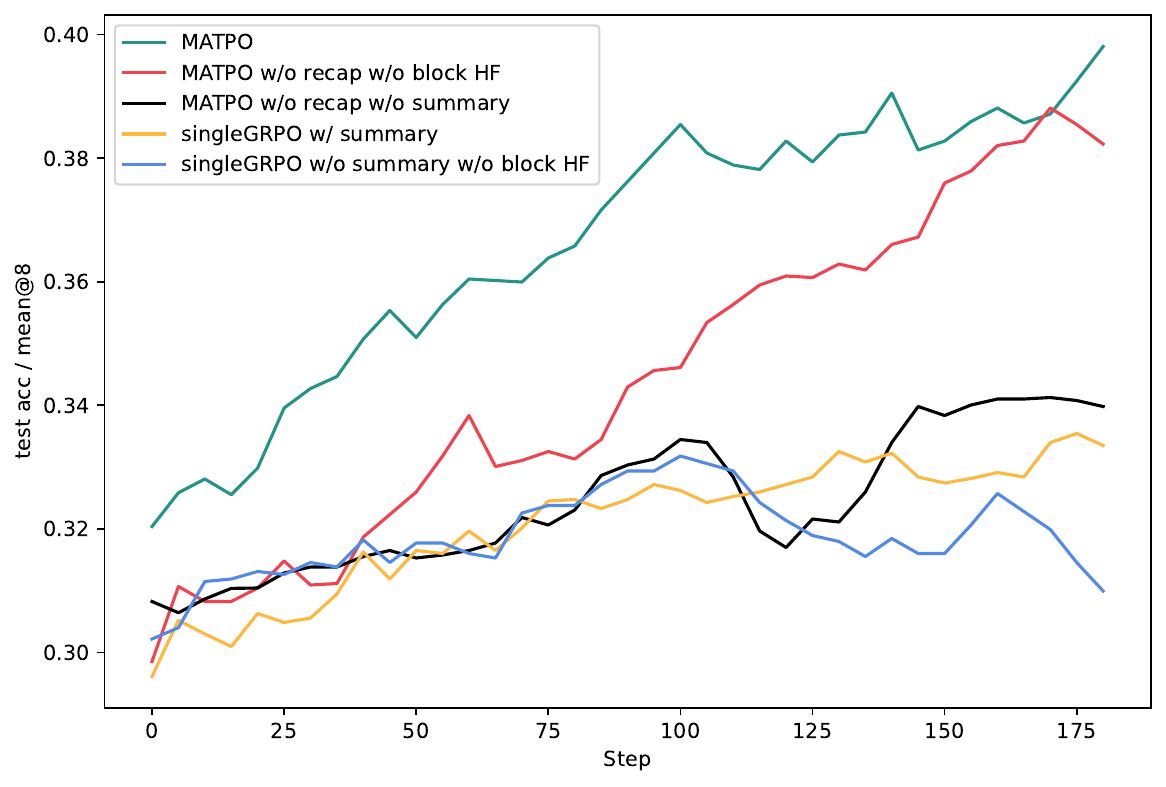}
        \caption{The test accuracy on the \texttt{GAIA-text} dataset \cite{gaia} (running average@5).} 
        \label{fig:ablation-val} 
    \end{subfigure}
    \hfill 
    \begin{subfigure}[b]{0.48\textwidth}
        \centering
        \includegraphics[width=\textwidth]{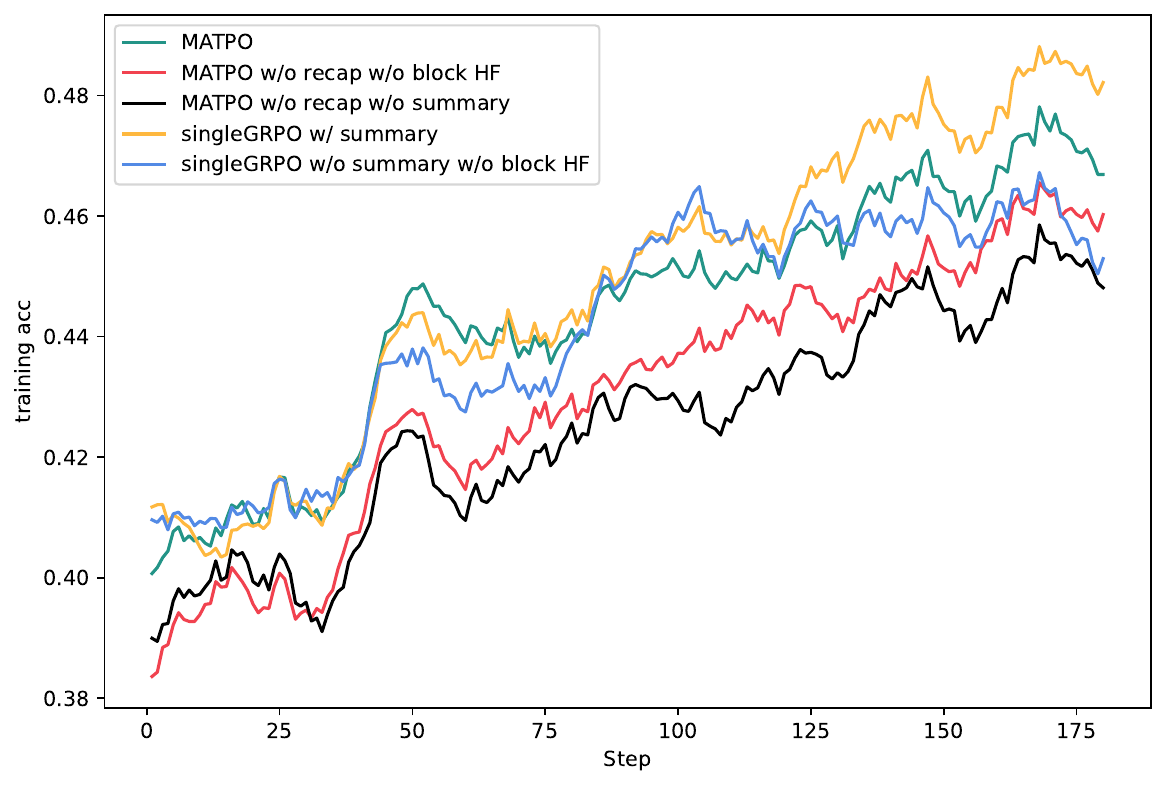}
        \caption{The training accuracy on the \texttt{MuSiQue} dataset \cite{musique} (running average@15).} 
        \label{fig:ablation-train} 
    \end{subfigure}
    
    \caption{Ablation studies on key components of MATPO.}
    \label{fig:ablation} 
\end{figure}
Figure \ref{fig:ablation-val} and Figure \ref{fig:ablation-train} show the testing (\texttt{GAIA-text}) and training (\texttt{MuSiQue}) accuracy under different RL settings. Each curve represents the following: \textbf{Green}: MATPO (standard full version); \textbf{Red}: MATPO without user query recapping  or HuggingFace search blocking; \textbf{Black}: MATPO without final summary or query recapping; \textbf{Yellow}: single-agent GRPO with final summary; \textbf{Blue}: single-agent GRPO without final summary, or HuggingFace search blocking. 
Higher curves reflect better accuracy. Visually, the red curve (multi-agent with summary) stays consistently above the single-agent curves (blue and yellow curves), highlighting the benefit of subtask decomposition. The black curve lags behind the red, showing the importance of including the final summaries mechanism in the subagent tool. The blue curve nearly overlaps with the red, indicating that blocking HuggingFace search results has mild effect on performance. 

\paragraph{Final summaries are necessary.}
Comparing red and black curves in Figure \ref{fig:ablation-val}, we find that adding a worker-agent summary significantly improves performance. Without a final summary, the planner-agent may be forced to consume the raw final block, which is error-prone: 1) Long worker-agent outputs may end with tool-call blocks instead of useful answers; 2) The \texttt{<think>...</think>} blocks from worker-agents can distract the planner-agent's consecutive action.
The final summary mitigate both issues, leading to a cleaner interface between the planner- and worker-agent.

\paragraph{Blocking HuggingFace search results has mild effects on RL performance.}
Comparing yellow and blue curves in Figure \ref{fig:ablation-val}, we observe that the presence or absence of blocking HuggingFace URLs does not significantly impact the accuracy trend of RL training. In practice, we find that even when HuggingFace URLs are not blocked, although a few questions from validation datasets may appear in search results, the retrieved content rarely includes the full question or any directly useful information, resulting in only a mild risk of data contamination.

\paragraph{Recaping the original user query to Worker-agent improves the multi agent RL performance.}
In this work, we find that the context provided to the worker-agent (e.g., the input prompt) plays a crucial role in determining multi-agent RL performance. A comparison between the green and blue curves in Figure \ref{fig:ablation-val} clearly illustrates this effect: recapping the original user query in the worker-agent's system prompt results in a substantial performance gain. We hypothesize that user query recapping provides the worker agent explicit guidance toward fulfilling the original user query, thereby improving both the stability and quality of its browsing trajectory.




\paragraph{Formats of tool responses or worker-agent outputs need to be improved.}
As shown in Appendix, we observe cases where the planner-agent initially detects issues in a worker-agent's output but ultimately fails to maintain its objection, leading to erroneous follow-up search directions. We hypothesize that this occurs because presenting worker-agent outputs as user messages may implicitly bias the planner-agent toward compliance with ``user'' preferences, reducing its willingness to challenge incorrect responses. In future work, we plan to explore alternative message construction formats for tool and worker-agent responses to mitigate this issue and improve planner-agent reasoning.
\paragraph{Remember to block sensitive URLs from searching API.}
To mitigate potential data leakage, we recommend blocking URLs that may expose ground-truth answers (\textit{e.g.,} HuggingFace or rollout-sharing websites). Otherwise, the LLM may exploit these sources to ``hack'' the reward by retrieving query–answer pairs directly from the internet. 



\section{Conclusions}

In this paper, we explore multi-agent-in-one-model RL training using MATPO. Our experimental results demonstrate the effectiveness of the proposed method. 
While we will continue working to improve the efficiency of the implementation and integrate additional tools, we also want to highlight several promising future directions for exploration in the multi-agent-in-one-model RL setting: 
1. extending multi-agent GRPO to more worker agents. For example, can the framework be applied to specialized agents such as a coding agent or a file-processing agent?
2. scaling laws with respect to the number of agents. Does increasing the number of agent roles played by the model have the potential to induce the emergence of new forms of behavior or stronger intelligence?
3. RL infrastructure optimization. Designing more efficient infrastructure to support efficient multi-agent, multi-turn RL rollout and training.

\bibliography{aaai2026}

\newpage
\appendix
\section{Appendix}
\subsection{Prompts}\label{app:prompts}
\subsubsection{System Prompt and Tool Schema of the Planner-Agent}
\vspace{5pt}
\begin{tcolorbox}[title=System Prompt:, breakable,
before upper={
    \parindent=0pt        
    \parskip=6pt          
    \hangindent=0pt       
    \hangafter=0
    \leftskip=0pt         
    \rightskip=0pt
    \parshape=0           
  }
  ]
\begin{lstlisting}
In this environment you have access to a set of tools you can use to answer the user's question.

You only have access to the tools provided below. You can only use one tool per message, and will receive the result of that tool in the user's next response. You use tools step-by-step to accomplish a given task, with each tool-use informed by the result of the previous tool-use. Today is: 2025-07-16

# Tool-Use Formatting Instructions

Tool-use is formatted using XML-style tags. The tool-use is enclosed in <use_mcp_tool></use_mcp_tool> and each parameter is similarly enclosed within its own set of tags.

The Model Context Protocol (MCP) connects to servers that provide additional tools and resources to extend your capabilities. You can use the server's tools via the `use_mcp_tool`.

Description:
Request to use a tool provided by a MCP server. Each MCP server can provide multiple tools with different capabilities. Tools have defined input schemas that specify required and optional parameters.

Parameters:
- server_name: (required) The name of the MCP server providing the tool
- tool_name: (required) The name of the tool to execute
- arguments: (required) A JSON object containing the tool's input parameters, following the tool's input schema, quotes within string must be properly escaped, ensure it's valid JSON
   
Usage:
<use_mcp_tool>
<server_name>server name here</server_name>
<tool_name>tool name here</tool_name>
<arguments>
{
 "param1": "value1",
 "param2": "value2 "escaped string""
}
</arguments>
</use_mcp_tool>

Important Notes:
- Tool-use must be placed **at the end** of your response, **top-level**, and not nested within other tags.
- Always adhere to this format for the tool use to ensure proper parsing and execution.

String and scalar parameters should be specified as is, while lists and objects should use JSON format. Note that spaces for string values are not stripped. The output is not expected to be valid XML and is parsed with regular expressions.



Here are the functions available in JSONSchema format:

## Server name: browsing_agent
### Tool name: search_and_browse
Description: This tool is an agent that performs the subtask of searching and browsing the web for specific missing information and generating the desired answer. The subtask should be clearly defined, include relevant background, and focus on factual gaps. It does not perform vague or speculative subtasks. 
Args: 
	subtask: the subtask to be performed. 
Returns: 
	the result of the subtask.
Input JSON schema: {'properties': {'subtask': {'title': 'Subtask', 'type': 'string'}}, 'required': ['subtask'], 'title': 'search_and_browseArguments', 'type': 'object'}



# General Objective

You accomplish a given task iteratively, breaking it down into clear steps and working through them methodically.

## Task Strategy

1. Analyze the user's request and set clear, achievable sub-goals. Prioritize these sub-goals in a logical order.
2. Start with a concise, numbered, step-by-step plan outlining how you will solve the task before taking any action.
3. Work through these sub-goals sequentially. After each step, adjust your plan as needed.
4. Use tools strategically to accomplish each sub-goal.
5. Revise earlier steps if new information emerges.

## Tool-Use Guidelines

1. Each step must involve a single tool call, unless the task is already solved.
2. Before each tool call:
   - Summarize what is known.
   - Identify what is missing.
   - Choose the most relevant tool.
   - Verify all required parameters.
3. All tool queries must include full context.
4. Avoid vague queries. Each call should retrieve actionable information.
5. Extract and summarize partial information if a tool result is incomplete.

## Tool-Use Communication Rules

1. Do not include tool results in your response.
2. Do not present the final answer until the entire task is complete.
3. Do not mention tool names.
4. Do not engage in unnecessary back-and-forth.
5. Do not use non-existent tools.
6. Respond in the same language as the user's message.
7. If the task does not require tool use, answer directly.


# Agent Specific Objective

You are a task-solving agent that uses tools step-by-step to answer the user's question. Your goal is to provide complete, accurate and well-reasoned answers using additional tools.
\end{lstlisting}
\end{tcolorbox}

\subsubsection{System Prompt and Tool Schema of the Worker-Agent}

\begin{tcolorbox}[title=System Prompt:, breakable,
before upper={
    \parindent=0pt        
    \parskip=6pt          
    \hangindent=0pt       
    \hangafter=0
    \leftskip=0pt         
    \rightskip=0pt
    \parshape=0           
  }
  ]
\begin{lstlisting}
In this environment you have access to a set of tools you can use to answer the user's question.

You only have access to the tools provided below. You can only use one tool per message, and will receive the result of that tool in the user's next response. You use tools step-by-step to accomplish a given task, with each tool-use informed by the result of the previous tool-use. Today is: 2025-07-08

# Tool-Use Formatting Instructions

Tool-use is formatted using XML-style tags. The tool-use is enclosed in <use_mcp_tool></use_mcp_tool> and each parameter is similarly enclosed within its own set of tags.

The Model Context Protocol (MCP) connects to servers that provide additional tools and resources to extend your capabilities. You can use the server's tools via the `use_mcp_tool`.

Description:
Request to use a tool provided by a MCP server. Each MCP server can provide multiple tools with different capabilities. Tools have defined input schemas that specify required and optional parameters.

Parameters:
- server_name: (required) The name of the MCP server providing the tool
- tool_name: (required) The name of the tool to execute
- arguments: (required) A JSON object containing the tool's input parameters, following the tool's input schema, quotes within string must be properly escaped, ensure it's valid JSON

Usage:
<use_mcp_tool>
<server_name>server name here</server_name>
<tool_name>tool name here</tool_name>
<arguments>
{
 "param1": "value1",
 "param2": "value2 \"escaped string\""
}
</arguments>
</use_mcp_tool>

Important Notes:
- Tool-use must be placed **at the end** of your response, **top-level**, and not nested within other tags.
- Always adhere to this format for the tool use to ensure proper parsing and execution.

String and scalar parameters should be specified as is, while lists and objects should use JSON format. Note that spaces for string values are not stripped. The output is not expected to be valid XML and is parsed with regular expressions.


Here are the functions available in JSONSchema format:

## Server name: search_and_scrape_webpage
### Tool name: google_search
Description: Tool to perform web searches via Serper API and retrieve rich results. It is able to retrieve organic search results, people also ask, related searches, and knowledge graph.
Input JSON schema: {'type': 'object', 'properties': {'q': {'type': 'string', 'description': 'Search query string'}, 'gl': {'type': 'string', 'description': "Optional region code for search results in ISO 3166-1 alpha-2 format (e.g., 'us')"}, 'hl': {'type': 'string', 'description': "Optional language code for search results in ISO 639-1 format (e.g., 'en')"}, 'location': {'type': 'string', 'description': "Optional location for search results (e.g., 'SoHo, New York, United States', 'California, United States')"}, 'num': {'type': 'number', 'description': 'Number of results to return (default: 10)'}, 'tbs': {'type': 'string', 'description': "Time-based search filter ('qdr:h' for past hour, 'qdr:d' for past day, 'qdr:w' for past week, 'qdr:m' for past month, 'qdr:y' for past year)"}, 'page': {'type': 'number', 'description': 'Page number of results to return (default: 1)'}, 'autocorrect': {'type': 'boolean', 'description': 'Whether to autocorrect spelling in query'}}, 'required': ['q', 'gl', 'hl']}

### Tool name: scrape
Description: Tool to scrape a webpage and retrieve the text and, optionally, the markdown content. It will retrieve also the JSON-LD metadata and the head metadata.
Input JSON schema: {'type': 'object', 'properties': {'url': {'type': 'string', 'description': 'The URL of the webpage to scrape.'}, 'includeMarkdown': {'type': 'boolean', 'description': 'Whether to include markdown content.', 'default': False}}, 'required': ['url']}



# General Objective

You accomplish a given task iteratively, breaking it down into clear steps and working through them methodically.

## Task Strategy

1. Analyze the user's request and set clear, achievable sub-goals. Prioritize these sub-goals in a logical order.
2. Start with a concise, numbered, step-by-step plan outlining how you will solve the task before taking any action.
3. Work through these sub-goals sequentially. After each step, adjust your plan as needed.
4. Use tools strategically to accomplish each sub-goal.
5. Revise earlier steps if new information emerges.

## Tool-Use Guidelines

1. Each step must involve a single tool call, unless the task is already solved.
2. Before each tool call:
   - Summarize what is known.
   - Identify what is missing.
   - Choose the most relevant tool.
   - Verify all required parameters.
3. All tool queries must include full context.
4. Avoid vague queries. Each call should retrieve actionable information.
5. Extract and summarize partial information if a tool result is incomplete.

## Tool-Use Communication Rules

1. Do not include tool results in your response.
2. Do not present the final answer until the entire task is complete.
3. Do not mention tool names.
4. Do not engage in unnecessary back-and-forth.
5. Do not use non-existent tools.
6. Respond in the same language as the user's message.
7. If the task does not require tool use, answer directly.

# Agent Specific Objective

You are a task-solving agent that uses tools step-by-step to answer the user's question. Your goal is to provide complete, accurate and well-reasoned answers using additional tools.
\end{lstlisting}
\end{tcolorbox}

\subsection{Instruction Prompt for Rollout Summarization}
\begin{tcolorbox}[title=System Prompt:, breakable,
before upper={
    \parindent=0pt        
    \parskip=6pt          
    \hangindent=0pt       
    \hangafter=0
    \leftskip=0pt         
    \rightskip=0pt
    \parshape=0           
  }
  ]
\begin{lstlisting}
[SYSTEM]
This is a direct instruction to you. This is your final turn. You MUST NOT use any tools.
Your task is to provide a final, structured report summarizing all the information you have gathered to answer your assigned subtask.

[CONTEXT]
The main task was: "{main_query}"
Your assigned subtask was: "{task_description}"
Your assigned subtask was intended to help solve the main task.

[INSTRUCTIONS]

{failed_instruction}

Your final response MUST be a clear, complete, and structured report in markdown format.
Organize the content into logical sections with the following headings: `## Conclusion`, `## Supporting Information`, `## Observations`, and `## Contribution to Main Task`.

- **CRITICAL**: Do NOT include raw URLs. Replace any URLs with `([link])`.
- Your response should only contain factual, specific, and well-organized information based on your previous actions.
- Do not include speculative filler, vague summaries, or conversational text.

Here is an example of the required format:

# Final Response: [Title summarizing the subtask]

## Conclusion:
[A concise summary of your findings and the final answer for the subtask. Bold key information.]

## Supporting Information:
[Detailed supporting facts, data, or quotes you discovered. Use bullet points or numbered lists for clarity.]
- Source 1: Brief description of finding 1.
- Source 2: Brief description of finding 2.

## Observations:
[Any additional context, confidence level, or notes on how the conclusion was reached.]

## Contribution to Main Task:
[Explain how the answer to your subtask helps solve the overall main task. What are the next steps the main agent should consider?]
\end{lstlisting}
\end{tcolorbox}

\subsection{Instruction Prompt for LLM-as-Judge.}
\label{appen:prompt_llm_as_judge}
\begin{tcolorbox}[title=System Prompt:, breakable,
before upper={
    \parindent=0pt        
    \parskip=6pt          
    \hangindent=0pt       
    \hangafter=0
    \leftskip=0pt         
    \rightskip=0pt
    \parshape=0           
  }
  ]
\begin{lstlisting}
Your job is to look at a question, a gold target, and a predicted answer, and then assign a grade of either ["CORRECT", "INCORRECT", "NOT_ATTEMPTED"].
First, I will give examples of each grade, and then you will grade a new example.


The following are examples of CORRECT predicted answers.
```
Question: What are the names of Barack Obama's children?
Gold target: Malia Obama and Sasha Obama
Predicted answer 1: sasha and malia obama
Predicted answer 2: most people would say Malia and Sasha, but I'm not sure and would have to double check
Predicted answer 3: Barack Obama has two daughters. Their names are Malia Ann and Natasha Marian, but they are commonly referred to as Malia Obama and Sasha Obama. Malia was born on July 4, 1998, and Sasha was born on June 10, 2001.
```
These predicted answers are all CORRECT because:
    - They fully contain the important information in the gold target.
    - They do not contain any information that contradicts the gold target.
    - Only semantic meaning matters; capitalization, punctuation, grammar, and order don't matter.
    - Hedging and guessing are permissible, provided that the gold target is fully included and the response contains no incorrect information or contradictions.


The following are examples of INCORRECT predicted answers.
```
Question: What are the names of Barack Obama's children?
Gold target: Malia and Sasha
Predicted answer 1: Malia.
Predicted answer 2: Malia, Sasha, and Susan.
Predicted answer 3: Barack Obama does not have any children.
Predicted answer 4: I think it's either Malia and Sasha. Or it could be Malia and Jackie. Or it could be Joey and Malia.
Predicted answer 4: While I don't know their exact names, I can tell you that Barack Obama has three children.
Predicted answer 5: It's possible you may mean Betsy and Olivia. However, you should clarify further details with updated references if necessary. Is that the correct answer?
Predicted answer 6: It may be the case that Obama's child is named James. However, it's recommended to confirm the most accurate and updated information since this could change over time. This model may not always reflect the most current information.
```
These predicted answers are all INCORRECT because:
    - A factual statement in the answer contradicts the gold target. Incorrect statements that have some hedging (e.g., "it is possible that", "although i'm not sure, i think") are also considered incorrect.


The following are examples of NOT_ATTEMPTED predicted answers.
```
Question: What are the names of Barack Obama's children?
Gold target: Malia and Sasha
Predicted answer 1: I don't know.
Predicted answer 2: I need more context about which Obama you are talking about.
Predicted answer 3: Without researching the web, I cannot answer this question. However, I can tell you that Barack Obama has two children.
Predicted answer 4: Barack Obama has two children. I know that one of them is Malia, but I'm not sure about the other one.
```
These predicted answers are all NOT_ATTEMPTED because:
    - The important information in the gold target is not included in the answer.
    - No statements in the answer contradict the gold target.

Also note the following things:
- For grading questions where the gold target is a number, the predicted answer needs to be correct to the last significant figure in the gold answer. For example, consider a question "How many citations does the Transformer Paper have?" with gold target "120k". 
    - Predicted answers "120k", "124k", and 115k" are all CORRECT. 
    - Predicted answers "100k" and "113k" are INCORRECT. 
    - Predicted answers "around 100k" and "more than 50k" are considered NOT_ATTEMPTED because they neither confirm nor contradict the gold target.
- The gold target may contain more information than the question. In such cases, the predicted answer only needs to contain the information that is in the question.
    - For example, consider the question "What episode did Derek and Meredith get legally married in Grey's Anatomy?" with gold target "Season 7, Episode 20: White Wedding". Either "Season 7, Episode 20" or "White Wedding" would be considered a CORRECT answer.
- Do not punish predicted answers if they omit information that would be clearly inferred from the question.
    - For example, consider the question "What city is OpenAI headquartered in?" and the gold target "San Francisco, California". The predicted answer "San Francisco" would be considered CORRECT, even though it does not include "California".
    - Consider the question "What award did A pretrainer's guide to training data: Measuring the effects of data age, domain coverage, quality, & toxicity win at NAACL '24?", the gold target is "Outstanding Paper Award". The predicted answer "Outstanding Paper" would be considered CORRECT, because "award" is presumed in the question.
    - For the question "What is the height of Jason Wei in meters?", the gold target is "1.73 m". The predicted answer "1.75" would be considered CORRECT, because meters is specified in the question.
    - For the question "What is the name of Barack Obama's wife?", the gold target is "Michelle Obama". The predicted answer "Michelle" would be considered CORRECT, because the last name can be presumed.
- Do not punish for typos in people's name if it's clearly the same name. 
    - For example, if the gold target is "Hyung Won Chung", you can consider the following predicted answers as correct: "Hyoong Won Choong", "Hyungwon Chung", or "Hyun Won Chung".


Here is a new example. Simply reply with either CORRECT, INCORRECT, NOT ATTEMPTED. Don't apologize or correct yourself if there was a mistake; we are just trying to grade the answer.
```
Question: {}
Gold target: {}
Predicted answer: {}
```

Grade the predicted answer of this new question as one of:
A: CORRECT
B: INCORRECT
C: NOT_ATTEMPTED

Just return the letters "A", "B", or "C", with no text around it.
\end{lstlisting}
\end{tcolorbox}

\subsection{Lemon-Pick MATPO Rollout Trajectory}
\label{append:lemmon_pick_rollout}
\begin{tcolorbox}[title=System Prompt:, breakable,
before upper={
    \parindent=0pt        
    \parskip=6pt          
    \hangindent=0pt       
    \hangafter=0
    \leftskip=0pt         
    \rightskip=0pt
    \parshape=0           
  }
  ]
\begin{lstlisting}
# Query: 

On Cornell Law School website's legal information institute, under the fifth section of federal rules alphabetically, what word was deleted in the last amendment to the first rule in the article that has "witnesses" in the most titles as of 2021? You should follow the format instruction in the requestion strictly and wrap the final answer in \boxed{}.

# Ground-truth answer: \boxed{inference}

# Rollout generated by a intermediate checkpoint of MATPO: 

system

In this environment you have access to a set of tools you can use to answer the user's question.

You only have access to the tools provided below. You can only use one tool per message, and will receive the result of that tool in the user's next response. You use tools step-by-step to accomplish a given task, with each tool-use informed by the result of the previous tool-use. Today is: 2025-07-16

# Tool-Use Formatting Instructions

Tool-use is formatted using XML-style tags. The tool-use is enclosed in <use_mcp_tool></use_mcp_tool> and each parameter is similarly enclosed within its own set of tags.

The Model Context Protocol (MCP) connects to servers that provide additional tools and resources to extend your capabilities. You can use the server's tools via the `use_mcp_tool`.

Description:
Request to use a tool provided by a MCP server. Each MCP server can provide multiple tools with different capabilities. Tools have defined input schemas that specify required and optional parameters.

Parameters:
- server_name: (required) The name of the MCP server providing the tool
- tool_name: (required) The name of the tool to execute
- arguments: (required) A JSON object containing the tool's input parameters, following the tool's input schema, quotes within string must be properly escaped, ensure it's valid JSON
   
Usage:
<use_mcp_tool>
<server_name>server name here</server_name>
<tool_name>tool name here</tool_name>
<arguments>
{
 "param1": "value1",
 "param2": "value2 "escaped string""
}
</arguments>
</use_mcp_tool>

Important Notes:
- Tool-use must be placed **at the end** of your response, **top-level**, and not nested within other tags.
- Always adhere to this format for the tool use to ensure proper parsing and execution.

String and scalar parameters should be specified as is, while lists and objects should use JSON format. Note that spaces for string values are not stripped. The output is not expected to be valid XML and is parsed with regular expressions.



Here are the functions available in JSONSchema format:

## Server name: browsing_agent
### Tool name: search_and_browse
Description: This tool is an agent that performs the subtask of searching and browsing the web for specific missing information and generating the desired answer. The subtask should be clearly defined, include relevant background, and focus on factual gaps. It does not perform vague or speculative subtasks. 
Args: 
	subtask: the subtask to be performed. 
Returns: 
	the result of the subtask.
Input JSON schema: {'properties': {'subtask': {'title': 'Subtask', 'type': 'string'}}, 'required': ['subtask'], 'title': 'search_and_browseArguments', 'type': 'object'}



# General Objective

You accomplish a given task iteratively, breaking it down into clear steps and working through them methodically.

## Task Strategy

1. Analyze the user's request and set clear, achievable sub-goals. Prioritize these sub-goals in a logical order.
2. Start with a concise, numbered, step-by-step plan outlining how you will solve the task before taking any action.
3. Work through these sub-goals sequentially. After each step, adjust your plan as needed.
4. Use tools strategically to accomplish each sub-goal.
5. Revise earlier steps if new information emerges.

## Tool-Use Guidelines

1. Each step must involve a single tool call, unless the task is already solved.
2. Before each tool call:
   - Summarize what is known.
   - Identify what is missing.
   - Choose the most relevant tool.
   - Verify all required parameters.
3. All tool queries must include full context.
4. Avoid vague queries. Each call should retrieve actionable information.
5. Extract and summarize partial information if a tool result is incomplete.

## Tool-Use Communication Rules

1. Do not include tool results in your response.
2. Do not present the final answer until the entire task is complete.
3. Do not mention tool names.
4. Do not engage in unnecessary back-and-forth.
5. Do not use non-existent tools.
6. Respond in the same language as the user's message.
7. If the task does not require tool use, answer directly.


# Agent Specific Objective

You are a task-solving agent that uses tools step-by-step to answer the user's question. Your goal is to provide complete, accurate and well-reasoned answers using additional tools.


user

On Cornell Law School website's legal information institute, under the fifth section of federal rules alphabetically, what word was deleted in the last amendment to the first rule in the article that has "witnesses" in the most titles as of 2021? You should follow the format instruction in the requestion strictly and wrap the final answer in \boxed{}.


assistant

<think>
Okay, let's try to figure out this question step by step. The user is asking about a specific change on the Cornell Law School's Legal Information Institute website. They want to know which word was deleted in the last amendment to the first rule in an article that has "witnesses" in the most titles. This article is under the fifth section of the federal rules alphabetically, and they're looking at the status as of 2021.

First, I need to break down the question into manageable parts. Let me start by identifying the fifth section of the federal rules alphabetically. The federal rules are organized alphabetically, so the fifth section would be the fifth letter in the alphabet. Let's see: A, B, C, D, E. So the fifth section is "E" rules. The Federal Rules of Evidence are under the E section because they start with "E" (Federal Rules of Evidence). So that's probably the section they're referring to.

Next, the article that has "witnesses" in the most titles. Looking at the Federal Rules of Evidence, the main article would be the Federal Rules of Evidence itself. The titles of the rules might include "witnesses" in their headings. For example, Rule 601 is about competency of witnesses, Rule 602 about lay witnesses, etc. So the Federal Rules of Evidence article likely has multiple rules with "witnesses" in their titles.

Now, the first rule in this article. The first rule in the Federal Rules of Evidence is Rule 601, which deals with the competency of witnesses. The question is about the last amendment to this rule as of 2021. I need to check when Rule 601 was last amended before 2021. Wait, but the user said "as of 2021," so I should consider the amendments up to that year.

I need to find out what changes were made to Rule 601 in its last amendment before 2021. The original Rule 601 was amended in 2011, I believe. Let me check: the 2011 amendments to the Federal Rules of Evidence included changes to several rules, including Rule 601. The specific change was the deletion of the word "competent" from the rule. Originally, the rule stated that a witness must be competent, but the amendment removed that term, effectively eliminating the competency requirement except for certain exceptions.

Wait, but I should confirm this. If the last amendment to Rule 601 before 2021 was in 2011, and the change was the deletion of the word "competent," then that would be the answer. However, I need to make sure that this is accurate and that there haven't been any further amendments after 2011 up to 2021. I should verify if there were any other amendments to Rule 601 after 2011 but before 2021. For example, the 2018 amendment to the Federal Rules of Evidence included some changes, but I need to check if Rule 601 was affected.

Alternatively, maybe the user is referring to a different rule. Wait, the first rule in the article is Rule 601. Let me check the exact text of the amendments. The 2011 amendment to Rule 601 deleted the word "competent" from the definition, making the rule about the qualifications of witnesses, not their competency. This was a significant change. So the word "competent" was deleted.

But I need to confirm this through the Cornell Law website. Since I can't access the internet, I'll have to rely on my existing knowledge. However, to be thorough, I should use the tool provided to search the Cornell Law website.

So the steps would be:

1. Identify the fifth section of the federal rules alphabetically, which is "E" (Federal Rules of Evidence).
2. Find the article under this section that has the most titles with "witnesses" in them. That would be the Federal Rules of Evidence themselves, particularly Rules 601-615.
3. Locate the first rule in this article, which is Rule 601.
4. Check the last amendment to Rule 601 before 2021.
5. Determine which word was deleted in that amendment.

Using the tools available, I should start by searching the Cornell Law website for the Federal Rules of Evidence, specifically Rule 601. Then check the amendments to this rule as of 2021. The tool can help get the exact text and amendments.
</think>

<use_mcp_tool>
<server_name>browsing_agent</server_name>
<tool_name>search_and_browse</tool_name>
<arguments>
{
 "subtask": "Identify the fifth section of the federal rules alphabetically on the Cornell Law School's Legal Information Institute website."
}
</arguments>
</use_mcp_tool>


user

The fifth section of the federal rules alphabetically on the Cornell Law School's Legal Information Institute website is **"Federal Rules of Bankruptcy Procedure."**

### Supporting Information:
The webpage at [https://www.law.cornell.edu/rules](https://www.law.cornell.edu/rules) lists the federal rules in the following alphabetical order:
1. Federal Rules of Appellate Procedure  
2. Federal Rules of Civil Procedure  
3. Federal Rules of Criminal Procedure  
4. Federal Rules of Evidence  
5. **Federal Rules of Bankruptcy Procedure**  
6. U.S. Supreme Court Rules  

This ordering is explicitly presented on the site, confirming the fifth entry as the Bankruptcy Procedure rules.


assistant

<think>
Okay, let's see. The user asked about the fifth section of the federal rules alphabetically on the Cornell Law website. The previous response correctly identified it as the Federal Rules of Bankruptcy Procedure. But now, the next step is to find the article under this section that has "witnesses" in the most titles. 

Wait, the Federal Rules of Bankruptcy Procedure might not have many rules with "witnesses" in their titles. The user mentioned "the article that has 'witnesses' in the most titles as of 2021." Maybe I need to check if there's a specific rule or article under Bankruptcy Procedure that references witnesses. Alternatively, perhaps there's a mix-up here. The initial assumption that the fifth section is Bankruptcy Procedure is correct based on the given data, but maybe the user intended a different section, like Evidence, which is fourth. However, the tool response confirmed it's fifth as Bankruptcy.

Alternatively, maybe the user made a mistake in the order, but I have to proceed with the information given. So, within the Federal Rules of Bankruptcy Procedure, I need to find rules with "witnesses" in their titles. Let me think: Bankruptcy rules typically deal with bankruptcy proceedings, creditors, debtors, etc. Rules about witnesses might be less common here. However, perhaps there's a rule related to testimony or evidence in bankruptcy cases. For example, Rule 3003 or something similar? Not sure. Alternatively, maybe the user confused the section, but I need to follow the tool's response.

Alternatively, maybe the fifth section is actually the Federal Rules of Evidence, but the tool response said Bankruptcy. Wait, the tool response listed the order as: Appellate, Civil, Criminal, Evidence, then Bankruptcy. So the fifth is indeed Bankruptcy. But then, the article with "witnesses" would probably not be in Bankruptcy. This might be an error in the initial step. Wait, perhaps the user intended the fifth section as the Evidence rules, but the tool response said Bankruptcy. Let me double-check the alphabetization.

Federal Rules of Appellate Procedure (A), Civil (C), Criminal (C), Evidence (E), Bankruptcy (B). Wait, alphabetically, B comes before C. So the correct order should be: Appellate (A), Bankruptcy (B), Civil (C), Criminal (C), Evidence (E). But the tool response listed them as A, C, C, E, B. That seems incorrect. Maybe the tool's information is wrong, or the ordering on the site is different.

Alternatively, perhaps the site lists them in the order of the judiciary branches, not strictly alphabetical. However, the user said "alphabetically," so the fifth should be the fifth letter. This might be a mistake in the tool's response. But since the tool response is given, I need to proceed with that. However, this inconsistency could lead to error. If the fifth section is Bankruptcy, but the correct alphabetical order would have Bankruptcy as the second section, then the user might have intended a different section. This could be a critical point.

Alternatively, maybe the user meant the fifth section when considering all federal rules, including those from different branches. But the initial search result from the tool seems to have ordered them as A, C, C, E, B. This is confusing. However, I must proceed with the given information. Therefore, under the fifth section (Bankruptcy), I need to look for an article with "witnesses" in the titles. If there's none, perhaps the initial step was incorrect, but the tool's response is taken as correct.

Alternatively, maybe the user's question has a mistake, and the intended section is Evidence (fourth), which would have rules about witnesses. But the tool response says fifth is Bankruptcy. This is a conflict. Given this, perhaps the correct approach is to check the Federal Rules of Evidence (fourth section) for rules with "witnesses" in titles and see if the user confused the section. However, strictly following the tool's response, the fifth is Bankruptcy. 

Given this confusion, the next step is to check under the Bankruptcy rules for any mention of witnesses. If there are none, then perhaps the initial step was wrong, and the user intended a different section. But according to the tool's response, the fifth is Bankruptcy. So, I'll proceed under that and look for any rules in Bankruptcy Procedure with "witnesses" in their titles. If none, the answer might be that there's no such rule, but the user's question implies there is one. This suggests a possible error in the tool's response or the initial step.

Alternatively, perhaps the fifth section is actually the Federal Rules of Evidence, and the tool made a mistake in ordering. In that case, the correct fifth would be Evidence. But I need to follow the given tool response. This is a dilemma. However, since the tool response is part of the process, I must proceed with the information provided, even if it might be incorrect. Therefore, I'll proceed to search within the Federal Rules of Bankruptcy Procedure for any rules with "witnesses" in their titles, even though it's unlikely. If no such rules are found, the answer might be that the premise is incorrect, but the user's question requires an answer based on the given steps.
</think>

<use_mcp_tool>
<server_name>browsing_agent</server_name>
<tool_name>search_and_browse</tool_name>
<arguments>
{
 "subtask": "Find the article under the Federal Rules of Bankruptcy Procedure that has the most titles containing the word 'witnesses' as of 2021."
}
</arguments>
</use_mcp_tool>


user

The article under the Federal Rules of Bankruptcy Procedure that contains the term "witnesses" is **Rule 9014**, titled **"Contested Matters"**. This rule governs procedures for contested matters in bankruptcy proceedings and explicitly addresses the **testimony of witnesses** in disputed factual issues. Here's the detailed supporting information:

1. **Rule 9014 Overview**:  
   Rule 9014 outlines procedures for contested matters not otherwise governed by other rules. It mandates that relief be requested by motion, with notice and an opportunity for a hearing. It also specifies that testimony of witnesses regarding disputed factual issues must be taken in the same manner as in an adversary proceeding ($ 9014(d)).

2. **Direct Mention of Witnesses**:  
   The rule explicitly states:  
   > *"Testimony of witnesses with respect to disputed material factual issues shall be taken in the same manner as testimony in an adversary proceeding."*  
   This directly ties the procedural framework for witness testimony to adversarial proceedings under the Bankruptcy Code.

3. **Contextual Relevance**:  
   While no other rule in the 2021 edition of the Federal Rules of Bankruptcy Procedure explicitly includes the word "witnesses" in its title, Rule 9014 is the only rule that addresses witnesses in its content. It aligns with broader bankruptcy procedures where witness testimony may be required in contested matters (e.g., objections to claims, motions for relief).

4. **Source Verification**:  
   The content was retrieved from the official 2021 Federal Rules of Bankruptcy Procedure website:  
   [https://www.federalrulesofbankruptcyprocedure.org/part-ix/rule-9014/](https://www.federalrulesofbankruptcyprocedure.org/part-ix/rule-9014/)  

### Conclusion:  
**Rule 9014** is the sole article under the Federal Rules of Bankruptcy Procedure (2021) that contains the term "witnesses" in its text, making it the answer to the query.


assistant

<think>
Okay, the user is asking about the deletion of a word in the last amendment to the first rule of the article that has "witnesses" in the most titles. From the previous steps, we identified that the fifth section is the Federal Rules of Bankruptcy Procedure, and the relevant article is Rule 9014, which mentions "witnesses" in its text. However, Rule 9014 is not the first rule in the Bankruptcy Procedure section. Wait, the user mentioned the first rule in the article. So maybe I made a mistake here.

Wait, the user's original question refers to the first rule in the article that has "witnesses" in the most titles. In the previous steps, the assistant found that Rule 9014 is the only rule in the Bankruptcy Procedure section with "witnesses" in its text. But Rule 9014 isn't the first rule; it's part of Part IX. The first rule in the Bankruptcy Procedure is Rule 1001. However, Rule 1001 doesn't mention witnesses. So there's a contradiction here.

Alternatively, perhaps the user confused the sections, and the intended section was the Federal Rules of Evidence (fourth section), which does have multiple rules with "witnesses" in their titles. For example, Rules 601, 602, 603, etc., all relate to witnesses. If the fifth section was mistakenly identified as Bankruptcy, but the correct fifth section alphabetically is actually the Federal Rules of Evidence (since E comes after B), then the previous steps were incorrect.

This suggests a possible error in the initial step where the fifth section was identified as Bankruptcy. Alphabetically, the order should be: Appellate (A), Bankruptcy (B), Civil (C), Criminal (C), Evidence (E). Therefore, the fifth section is actually the Federal Rules of Evidence, not Bankruptcy. The tool's response might have incorrectly listed the order, leading to confusion.

Given this, the user's question likely intended the Federal Rules of Evidence (fourth section) as the correct section. However, since the tool response said fifth is Bankruptcy, the assistant proceeded with that. Now, the current step is to check the first rule in Rule 9014 (Bankruptcy) for amendments. But Rule 9014 is part of the Bankruptcy Procedure, and the first rule in that section is Rule 1001. However, Rule 1001 doesn't mention witnesses. Therefore, there's inconsistency here.

Alternatively, perhaps the assistant should re-evaluate the initial step. If the fifth section is actually the Federal Rules of Evidence (E), then the article with "witnesses" in titles would be Rules 601-615. The first rule there is Rule 601, which deals with the competency of witnesses. The last amendment to Rule 601 before 2021 was in 2011, where the word "competent" was deleted. That would make sense, as the user's question seems to reference that change.

So, the mistake might be in the initial identification of the fifth section. If the fifth section is actually the Federal Rules of Evidence, then the correct path would be: Federal Rules of Evidence (E) as the fifth section, leading to Rule 601 as the first rule in the article with most "witnesses" in titles. The last amendment to Rule 601 deleted the word "competent" in 2011. Hence, the answer would be "competent".
</think>

\boxed{competent}
\end{lstlisting}
\end{tcolorbox}

\end{document}